\pdfoutput=1
\documentclass{article}
\usepackage{iclr2027_conference,times}
\iclrfinalcopy

\usepackage{amsmath,amsfonts,bm}

\def\eqref#1{equation~\ref{#1}}

\def\1{\bm{1}}

\DeclareMathAlphabet{\mathsfit}{\encodingdefault}{\sfdefault}{m}{sl}
\SetMathAlphabet{\mathsfit}{bold}{\encodingdefault}{\sfdefault}{bx}{n}

\usepackage[utf8]{inputenc}
\usepackage[T1]{fontenc}
\usepackage{hyperref}
\hypersetup{colorlinks=true, linkcolor=red!75!black, citecolor=green!45!black, urlcolor=red!75!black}
\usepackage{url}
\usepackage{booktabs}
\usepackage{amsmath,amssymb}
\usepackage{microtype}
\usepackage{xcolor}
\usepackage{listings}
\usepackage{graphicx}
\usepackage{multirow}
\usepackage{subcaption}
\usepackage{tikz}
\usepackage{wrapfig}
\usepackage{enumitem}
\usepackage{titletoc}
\usepackage{float}
\floatstyle{ruled}
\newfloat{algorithm}{tbp}{loa}
\floatname{algorithm}{Algorithm}

\usetikzlibrary{positioning,arrows.meta,calc}

\newcommand{\dataurl}{https://huggingface.co/datasets/leanpolish-anon/lean-proof-compression}
\newcommand{\LP}{\textsc{LeanPolish}}

\definecolor{accent}{HTML}{3B5BDB}
\definecolor{accentlight}{HTML}{E7ECFB}
\definecolor{accepted}{HTML}{C7E9D3}
\definecolor{accepteddark}{HTML}{2F855A}
\definecolor{rejectfill}{HTML}{FCE8E4}
\definecolor{rejectdark}{HTML}{B23B25}
\definecolor{ink}{HTML}{1F2937}
\definecolor{inkmid}{HTML}{4B5563}
\definecolor{codebg}{HTML}{F8FAFC}
\definecolor{diffaddline}{HTML}{1E8E3E}
\definecolor{diffremline}{HTML}{B23B25}
\definecolor{diffctxbg}{HTML}{F8FAFC}
\definecolor{difflabelbg}{HTML}{EEF2F7}
\definecolor{difflabelfg}{HTML}{334155}

\lstdefinestyle{leandiff}{%
  basicstyle=\footnotesize\ttfamily, breaklines=true, breakatwhitespace=true, columns=fullflexible,
  keepspaces=true, showstringspaces=false, frame=none,
  xleftmargin=8pt, xrightmargin=8pt, aboveskip=4pt, belowskip=4pt,
  backgroundcolor=\color{diffctxbg},
  escapeinside={(*@}{@*)},
  moredelim=[l][\color{diffaddline}\bfseries]{+},
  moredelim=[l][\color{diffremline}\bfseries]{-},
}
\newcommand{\diffheader}[2]{%
  \par\nobreak\vspace{2pt}%
  \noindent\colorbox{difflabelbg}{%
    \parbox{\dimexpr\linewidth-2\fboxsep}{%
      \footnotesize\sffamily\color{difflabelfg}\textbf{#1}\hfill\texttt{#2}}}%
  \par\nobreak\vspace{-2pt}}

\title{LeanPolish: Verified Supervision for\\Lean Proof Compression}

\author{Pauline Bourigault \\
Imperial College London \thanks{Code: \href{https://github.com/paulinebourigault/leanpolish}{https://github.com/paulinebourigault/leanpolish}. Dataset: \href{https://huggingface.co/datasets/leanpolish-anon/lean-proof-compression}{https://huggingface.co/datasets/leanpolish-anon/lean-proof-compression}.}}

\begin{document}

\maketitle

\begin{abstract}
Verified proof edits offer a natural source of supervision for improving
language-model-generated Lean proofs. Yet verification establishes that
an edit is correct, not that its training signal is free of search
artifacts. We introduce \LP{}, a symbolic Lean~4 pipeline that releases
33{,}402 accepted local edits and 65{,}596 same-state failed attempts,
and use it to study what models learn from this supervision.
First-success search admits a goal-independent rule with perfect ranking
accuracy; teacher-selected evaluation sites also reward trivial
deletions. Continuing menu evaluation beyond the first success removes the ordering shortcut:
a trained ranker selects the best candidate on 70.1\% of evaluated
held-out states, versus 36.9\% for the strongest frozen baseline.
For compression, iterating the symbolic pass raises miniF2F savings
from 19.7\% to 27.5\%, exceeding the neural hybrids we test there.
Verified neural editing helps on other proof sources, but matched
frozen-model controls show that its gains need not come from training.
The supervision does improve whole-proof rewriting: fine-tuning raises
verified token reduction from 2.8\% to 5.5\% on 19 PutnamBench proofs.
Together, the released edits, complete candidate pools, and controlled
evaluations separate learning to imitate a search policy from improving
on that search. They provide a reproducible basis for studying proof
improvement while keeping correctness, compression, and edit policy distinct.

\end{abstract}


\section{Introduction}
\label{sec:intro}

Language-model provers now produce kernel-checked Lean~4
\citep{demoura2021lean4} proofs for competition problems at scale
\citep{ren2025deepseekv2, lin2025goedelv2, wang2025kimina,
chen2025seedprover15, achim2025aristotle, axiom2025putnam}. Correctness
does not make these proofs concise: search can leave speculative tactic
cascades, unused facts, and repeated derivations. Recent systems
therefore learn to shorten proofs using verified outputs of search
\citep{gu2025proofoptimizer, ahuja2026improver2, lu2026leanrefactor}.
Such supervision has an appealing intuition: a complete proof shows
that one derivation works, whereas a verified edit shows how to improve
that derivation in a specific context.

The difficulty is that the search also determines which contexts,
alternatives, and labels enter the dataset. A kernel certificate answers
\emph{``does this edit preserve the theorem?''}; it does not answer
\emph{``what must a model learn to predict this edit?''} If search stops
at its first success, its logged failures reveal the winner's position
in the search order. If evaluation uses only sites where the search
found an edit, it measures conditional imitation rather than autonomous
proof improvement. These effects can make a verified dataset look more
informative than its evaluation establishes.

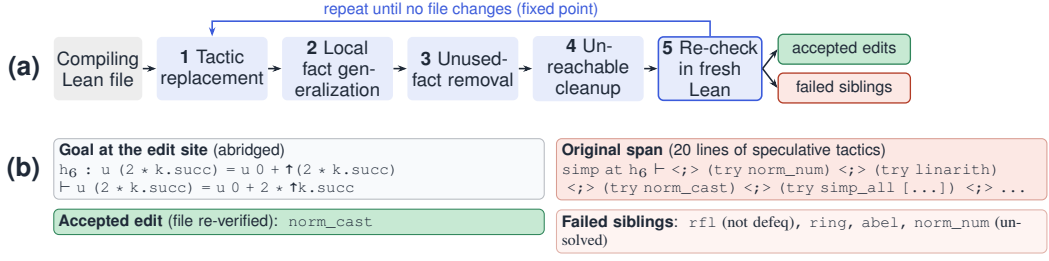
\begin{figure}[t]
\centering
\resizebox{\linewidth}{!}{%
\begin{tikzpicture}[font=\scriptsize\sffamily, every node/.style={text=ink},
  >={Stealth[length=3.5pt,width=2.5pt]},
  ph/.style={rectangle, rounded corners=2pt, fill=accentlight, minimum height=8.5mm, text width=14mm, align=center, inner sep=1.2pt},
  io/.style={rectangle, rounded corners=2pt, fill=black!7, minimum height=8.5mm, text width=11mm, align=center, inner sep=1.2pt},
  ver/.style={rectangle, rounded corners=2pt, draw=accent, line width=0.6pt, fill=accentlight, minimum height=8.5mm, text width=13mm, align=center, inner sep=1.2pt},
  outb/.style={rectangle, rounded corners=2pt, line width=0.5pt, minimum height=4mm, text width=17mm, align=center, inner sep=1pt, font=\tiny\sffamily},
  box/.style={rectangle, rounded corners=2pt, inner sep=2pt, align=left, text width=#1, font=\tiny\sffamily},
  flow/.style={->, line width=0.6pt, draw=ink}]
\node[io] (in) {Compiling\\Lean file};
\node[ph, right=1.8mm of in] (p1) {\textbf{1} Tactic replacement};
\node[ph, right=1.8mm of p1] (p2) {\textbf{2} Local fact generalization};
\node[ph, right=1.8mm of p2] (p3) {\textbf{3} Unused-fact removal};
\node[ph, right=1.8mm of p3] (p4) {\textbf{4} Unreachable cleanup};
\node[ver, right=1.8mm of p4] (p5) {\textbf{5} Re-check in fresh Lean};
\foreach \a/\b in {in/p1,p1/p2,p2/p3,p3/p4,p4/p5} \draw[flow] (\a) -- (\b);
\draw[flow, draw=accent] (p5.north) -- ++(0,2mm) -| node[pos=0.25, above, font=\tiny\sffamily, text=accent, inner sep=1pt] {repeat until no file changes (fixed point)} (p1.north);
\node[outb, draw=accepteddark, fill=accepted, right=2mm of p5, yshift=2.6mm] (acc) {accepted edits};
\node[outb, draw=rejectdark, fill=rejectfill, right=2mm of p5, yshift=-2.6mm] (rej) {failed siblings};
\draw[flow] (p5.east) -- (acc.west); \draw[flow] (p5.east) -- (rej.west);
\node[font=\bfseries\sffamily, anchor=east] at ($(in.west)+(-0.5mm,0)$) {(a)};
\node[box=0.46\linewidth, fill=codebg, draw=inkmid!40, below=5mm of in.south west, anchor=north west] (goal) {%
\textbf{Goal at the edit site} (abridged)\\
\ttfamily h$_6$ : u (2 * k.succ) = u 0 + \textuparrow(2 * k.succ)\\
\ttfamily $\vdash$ u (2 * k.succ) = u 0 + 2 * \textuparrow k.succ};
\node[box=0.46\linewidth, fill=rejectfill, draw=rejectdark!50, right=1.5mm of goal.north east, anchor=north west] (orig) {%
\textbf{Original span} (20 lines of speculative tactics)\\
\ttfamily simp at h$_6$ $\vdash$ <;> (try norm\_num) <;> (try linarith)\\
\ttfamily \ <;> (try norm\_cast) <;> (try simp\_all [...]) <;> ...};
\node[box=0.46\linewidth, fill=accepted, draw=accepteddark, below=1.2mm of goal] (accr) {%
\textbf{Accepted edit} (file re-verified):\ \ \ttfamily norm\_cast};
\node[box=0.46\linewidth, fill=rejectfill!50, draw=rejectdark!40, below=1.2mm of orig] (rejr) {%
\textbf{Failed siblings}:\ \ \ttfamily rfl\textrm{ (not defeq)}, ring, abel, norm\_num\textrm{ (unsolved)}};
\node[font=\bfseries\sffamily, anchor=east] at ($(goal.west)+(-0.5mm,0)$) {(b)};
\end{tikzpicture}
}
\caption{(a) The \LP{} pipeline: candidate tactics are tested at the
saved proof state, edited files are re-checked, and fixed-point iteration
re-applies the pipeline to its output. Complete-menu mode records all
menu outcomes. (b) One released record (miniF2F \texttt{aime\_1984\_p1}): a
learner sees the goal and the original span and must produce the edit;
the failed siblings are stored with it.}
\label{fig:overview}
\end{figure}

We study this distinction with \LP{}, a symbolic pipeline that replaces
goal-closing tactics, removes unused and unreachable material, and
factors repeated local facts (Figure~\ref{fig:overview}, Algorithm~\ref{alg:leanpolish}). Its central
output is a replayable local record: an original span, a verified
replacement, the available proof state, and alternatives tried at that
state. This granularity makes the teacher's decisions inspectable. We
use it to ask three questions: \emph{what signal do the records contain,
how should that signal be evaluated, and when does learning from it help
beyond running the symbolic search?}

The controls change the interpretation of apparently strong results.
A ranker reaches 100\% on first-success groups, but so does a rule that
ignores the goal and picks the latest candidate in the fixed menu. A
fine-tuned 7B editor recovers 92--98\% of the teacher's savings at
teacher-selected sites, yet 40\% of these sites are deletions identified
by a trivial prompt marker. On tactic-replacement sites, fine-tuning
reaches 42--79\% success, versus 0--18\% with four-shot prompting.
Recording complete menu outcomes removes the ordering shortcut;
on the resulting evaluation pools, a trained ranker reaches 70.1\%
top-1 accuracy, compared with 51.0\% without the goal and 36.9\% for
the strongest frozen baseline.

Learning this signal and improving a complete proof are different
tests. Iterating \LP{} to a fixed point raises miniF2F token reduction
from 19.7\% to 27.5\%, exceeding all tested neural hybrids on that
source. Neural proposals add on PutnamBench and AxiomProver, but a
matched frozen editor often achieves as much as a trained one, and
some trained edits relax the teacher's specificity policy. The clearest
training benefit is whole-proof rewriting: \LP{} proof pairs raise
verified reduction from 2.8\% to 5.5\% on PutnamBench. In short, the
kernel decides which edits are correct, but the search decides which
edits are recorded; \LP{} makes that second decision explicit, so it can
be controlled, removed, and tested.

\paragraph{Contributions.}
\begin{enumerate}[leftmargin=*]
\item \textbf{A neurosymbolic method for verified proof compression:}
  a symbolic Lean~4 pass whose every edit is kernel-checked and logged with
  its proof state and failed alternatives; a complete-menu mode that labels
  every candidate; and a verified hybrid (Alg.~\ref{alg:hybrid}) in which
  neural proposals are admitted only after joint re-verification
  (\S\ref{sec:method}, \S\ref{sec:use}). Its release comprises 33{,}402
  accepted edits, 65{,}596 linked failures, and complete pools for about
  60{,}000 proof states.
\item \textbf{An evaluation protocol that exposes search selection effects:}
  menu-order, deletion, prompting, and goal-ablation controls distinguish
  shortcuts from learnable state-dependent signal (\S\ref{sec:selection}).
\item \textbf{A controlled study of downstream utility:} whole-file
  verification, symbolic fixed points, frozen editors, and policy-matched
  comparisons distinguish gains from supervision from gains due to
  additional search or a different edit policy (\S\ref{sec:use}).
\end{enumerate}
Together, these contributions make the symbolic teacher, its training
signal, and its downstream utility independently inspectable.

\section{Related Work}
\label{sec:related}

\paragraph{Proof optimization.}
Lean provers trained on binary success \citep{polu2023curriculum, lample2022htps,
xin2024deepseekv15, ren2025deepseekv2, lin2025goedelv2, wang2025kimina,
chen2025seedprover15, achim2025aristotle, hubert2025alphaproof} produce long
proofs, and several systems now shorten them. ImProver \citep{ahuja2025improver},
Lean Refactor \citep{lu2026leanrefactor} and Proof-Refactor
\citep{fu2026proofrefactor} prompt frozen LLM agents; Lean Refactor retrieves
strategies distilled from 200K verified long--short pairs.
ProofOptimizer \citep{gu2025proofoptimizer} and ImProver~2 \citep{ahuja2026improver2}
train 7B models by expert iteration, with RL and preference optimization,
respectively, to rewrite \emph{whole} proofs. ProofOptimizer reports
87.9\%/57.2\% reductions on its Goedel-Prover-V2 miniF2F/PutnamBench
inputs; these use different proof samples and aggregation from ours
(App.~\ref{app:symbolic-main}). On the symbolic side, Mathlib's
linters, \texttt{simp?}/\texttt{hint}, and AXLE's \texttt{simplify\_theorems}
\citep{xin2026axle}, which removes unused tactics and \texttt{have}s to a fixed
point, overlap with our deletion phases. What differs here is the output and
the granularity: \LP{} adds goal-closing tactic replacement under a
specificity filter and local-fact anti-unification \citep{plotkin1970,
cerna2023antiunification}, and emits each accepted change as a
\emph{local, goal-conditioned} edit with the same-state failures, so that a
model is trained on a symbolic teacher's edits rather than on its own
whole-proof samples. These systems build optimizers; we study the
verified supervision such optimizers are trained on. Their expert
iteration and winner/loser pairs are search-generated data of the kind
whose selection effects we measure, and our controls (iterated symbolic
baseline, matched frozen models, policy compliance) separate gains due to
training from gains due to verification and search. The approaches are
complementary: \LP{} can pre-process their inputs (12.5\%\,$\to$\,27.6\%
for a frozen rewriter on miniF2F-99) and supply complete candidate pools.

\paragraph{Proof repair and local edits.}
Repair conditions on a failed proof and its error
\citep{first2023baldur, ospanov2025apollo, lin2025goedelv2}; APRIL
\citep{wang2026april} releases 260K state-conditioned repair tuples from
synthetic mutations, and BlueprintRepair \citep{khrulev2026blueprintrepair}
restricts repair to typed local operations with accepted and rejected
trajectories. ProofAug \citep{liu2025proofaug} tries automation at every
sub-proof of an LLM proof, as our tactic menu does, but to find proofs rather
than shorten them. Our edits act on \emph{correct} proofs, and our negatives
are natural failures of real automation on real goals, not mutations; our
hybrid (\S\ref{sec:hybrid}) reuses the verify-splice-recombine loop of APOLLO
for compression.

\paragraph{Learning from search and from failures.}
Training on the verified output of search is expert iteration
\citep{polu2023curriculum, lample2022htps}; using compiler-rejected tactics as
negatives appears in BFS-Prover's state--tactic DPO \citep{xin2025bfsprover},
trial-and-error fine-tuning \citep{an2024learnfromfailure}, tactic-level
verified rewards \citep{kim2026processverified}, and ImProver~2's
winner/loser rewrites. Our contribution on this axis is an empirical diagnosis: we show that negatives recorded in
search order admit a trivial ranking shortcut (\S\ref{sec:ranking}), and
provide a complete-menu mode that removes it. The closest analogy outside
theorem proving is learned compiler optimization, where models are trained on
the output of symbolic/autotuning search \citep{cummins2023llmopt,
cummins2025llmcompiler} or learn proposals for a verifier-gated rewrite search
\citep{schkufza2013stoke, bunel2017superopt, shypula2024pie}; \LP{} instantiates
this teacher-then-verified-student pattern for Lean proofs.

\paragraph{Lean datasets.}
LeanDojo \citep{yang2023leandojo}, Lean Workbook \citep{ying2024leanworkbook},
Goedel-Pset \citep{lin2025goedelv1}, Herald \citep{gao2025herald} and
FormalMATH \citep{yu2025formalmath} provide statements, proofs, or
traces. \LP{} instead aligns local spans and their verified replacements
with available goal states and same-state attempted alternatives. This
record structure supports both edit generation and controlled candidate
ranking.

\section{Generating Verified Edit Supervision}
\label{sec:method}

\LP{} turns a compiling proof into verified local improvement examples.
It uses Lean's \texttt{InfoTree}, which records the proof states produced
while elaborating the original file. Candidate tactics can therefore be
tested at the original state before re-checking the edited file. Four
phases run in order: tactic replacement, local fact generalization,
unused-fact removal, and unreachable-code cleanup. The key separation is
between \emph{proposal} (what to try), \emph{policy} (which edits to prefer),
and \emph{verification} (whether the resulting proof still checks).

\paragraph{Tactic replacement.}
At every leaf tactic that closes a goal, \LP{} tries a fixed menu in
order (\texttt{rfl}, \texttt{ring}, \texttt{abel}, \texttt{norm\_num},
\texttt{norm\_cast}, \texttt{positivity}, \texttt{decide},
\texttt{linarith}, \texttt{omega}, \texttt{field\_simp},
\texttt{contradiction}, \texttt{ext}, \texttt{gcongr}, \texttt{tauto},
\texttt{simp?} squeezed to \texttt{simp only [...]}), skipping any
candidate that is not shorter than the original, with 5\,s timeouts on
the expensive tactics. The \emph{first} candidate that closes the goal
is proposed; if none does, a bounded \texttt{exact?} call is tried. A
proposal is kept only if it passes a \emph{specificity filter}: tactics
that carry explicit terms, lemma names, or case structure supplied by
the proof author (\texttt{exact}, \texttt{rw}, \texttt{apply},
\texttt{refine}, \texttt{use}, \texttt{cases}, \texttt{induction},
\texttt{calc}, \texttt{conv}, \ldots) are never replaced unless the
original is a multi-step block, and a decision procedure is never
replaced by a more general one (e.g.\ \texttt{ring} only by
\texttt{rfl}, \texttt{norm\_num} never by \texttt{simp}). The filter encodes a maintainability preference;
it is not a correctness mechanism (App.~\ref{app:method}).

\paragraph{Local fact generalization.}
Within one proof, near-duplicate \texttt{have} blocks are grouped and
anti-unified; the differing subterms become parameters of one shared
local fact, which is kernel-checked and substituted at each use. A
merge is kept only if it saves bytes net of its own declaration and
passes a \emph{dependency filter}: the abstracted proof must depend on
strictly fewer local free variables than the union of the original
proofs, so that the parameters actually absorb local dependencies. This
is a heuristic against merges that merely re-package the originals; it
is not a guarantee of useful generalization, and its ablation leaves aggregate savings unchanged on the tested
slices (\S\ref{sec:ablation-summary}). The
filter applies to kernel-level extraction; exact duplicates and
same-statement \texttt{have} blocks are instead replaced by a reference
to the earlier hypothesis.

\paragraph{Unused facts and unreachable code.}
A usage graph identifies \texttt{have} blocks whose names are never
referenced later; they are removed together, falling back to one at a
time, and every removal is checked. When a file contains \texttt{<;>}
chains, it is re-elaborated with Lean's default linters (up to three
rounds) and tactics reported as never executed or as doing nothing are
removed from those chains.

\paragraph{Verification.}
Tactic candidates are tested against the saved proof state; the fully
edited text is re-elaborated and kernel-checked in-process before it is
saved. For the benchmark and competition sources (miniF2F,
PutnamBench, Putnam~2025, and the frontier releases of
\S\ref{sec:symbolic}), the saved file is additionally compiled by an
independent \texttt{lake env lean} process from a fresh Mathlib import;
if joint verification fails, a compiling subset of the edits is kept.
Independently of
the pipeline, the released verifier splices a
single edit into its source file and compiles it in a fresh process;
every model output counted as a success in \S\ref{sec:learning} was
checked this way. Edits touch only proof bodies, so theorem statements
are unchanged. We do not claim that outputs are length-minimal or more
readable to humans. The default toolchain is Lean~4.21.0 / Mathlib v4.21.0; frontier-source
exceptions are specified in App.~\ref{app:frontier}.

\paragraph{Complete-menu search.}
Stopping at the first success leaves later alternatives unobserved.
In \emph{complete-menu} mode, \LP{} continues through all 15 menu entries
and selects the shortest successful candidate allowed by the specificity
filter, breaking ties by menu order. It records the selected candidate,
other valid candidates, Lean failures, timeouts, policy rejections, and
candidates skipped for length separately. These are complete
\emph{search outcomes}; a timeout is not a proof of invalidity, and a
policy rejection is not a correctness failure. The observed pools have
2.2--3.4 valid candidates per site, with at least two at 55--74\% of
sites, at 1.1--1.3$\times$ the menu-search time. Choosing a different
one-token tactic can save characters without changing token counts;
we observe no additional token compression from this mode. Its purpose
is to remove first-success censoring, while retaining the declared menu
and edit policy as part of the task (\S\ref{sec:ranking}).

\paragraph{Fixed-point iteration.}
A single pass is not a fixed point: one edit can enable another (a
tactic replacement can leave an earlier fact unused), and per-file time
budgets cut some searches short. We therefore also run \LP{} on its own
output until no file changes (4--8 rounds, every round re-verified in a
fresh process). This exposes compression that a one-pass baseline leaves available (\S\ref{sec:use}).

\paragraph{What is recorded, and what the failures mean.}
Each accepted edit becomes a row with the available goal state
(three renderings for tactic edits),
the exact original and replacement spans with byte offsets, the edit
kind, token/byte/line counts, and provenance (source file, toolchain
and optimizer revisions). Each candidate that failed on the same goal
in the same attempt becomes a linked \emph{rejected sibling} with its
error message. Because the menu is searched in a fixed order and stops
at the first success, a sibling means ``tried before the winner and
failed'' (an error or a timeout); candidates after the winner were never
tried and are \emph{unknown}, not negative. Siblings are recorded only
for accepted edits; proposals rejected by the specificity filter and
menu candidates skipped for not being shorter are not recorded. This
matters for how the negatives can be used (\S\ref{sec:ranking}).

\subsection{Released records}
\label{sec:resource}

\begin{table}[t]
\caption{Released accepted edits and same-attempt failures per source.
Token reduction is whole-corpus under the Lean-aware tokenizer
(App.~\ref{app:symbolic-main}), with files that are not shortened counted as
0\%. Files = files with at least one accepted edit. Edit density and mean savings per edit are given in
App.~\ref{app:symbolic-main}.}
\label{tab:corpora}
\centering
\small
\resizebox{\linewidth}{!}{%
\begin{tabular}{lrrrrr}
\toprule
Source (generator) & Inputs & Files & Accepted & Failed siblings & Tok.\ red.\ (\%) \\
\midrule
Mathlib v4.21.0 subset (human) & 5{,}789 & 2{,}233 & 6{,}695 & 26{,}912 & 0.27 \\
Goedel-Workbook (Goedel-Prover-V2) & 29{,}750 & 10{,}052 & 20{,}822 & 28{,}525 & 5.48 \\
PutnamBench sample (Goedel-Prover-V2) & 437 & 352 & 4{,}354 & 5{,}930 & 8.14 \\
miniF2F verified (Goedel-Prover-V2) & 351 & 308 & 1{,}184 & 3{,}753 & 19.72 \\
PutnamBench verified (Goedel-Prover-V2) & 19 & 16 & 80 & 254 & 6.14 \\
Putnam 2025 (AxiomProver) & 12 & 10 & 142 / 125 & 147 / 75 & 1.31 \\
\midrule
Total &  & 12{,}972$^\ddagger$ & 33{,}402 & 65{,}596 &  \\
\bottomrule
\end{tabular}}
\par\smallskip{\raggedright\footnotesize Putnam 2025 is processed under two scheduler
configurations (sequential / pooled), shipped as separate shards and
union-counted once in the file total; the displayed reduction uses the
sequential configuration. $^\ddagger$Distinct files with an accepted edit,
including 92 linter-baseline files (\texttt{*\_linter.lean}) listed in
the release; 12{,}880 are genuine inputs (App.~\ref{app:errata}).
\par}
\end{table}

Table~\ref{tab:corpora} summarizes the release (public dataset;
link in the Code and Data paragraph). About 36k
files were processed; 12{,}972 received at least one accepted edit. The five
sources span human-curated Mathlib, a large LLM-generated workbook
(Goedel-Workbook proofs sampled from Goedel-Prover-V2-32B
\citep{lin2025goedelv1, lin2025goedelv2}), two Goedel-Prover-V2 proof
pools on benchmark statements \citep{zheng2022minif2f,
tsoukalas2024putnambench}, and AxiomProver's Putnam~2025 solutions
\citep{axiom2025putnam}. By edit family, the 33{,}286 distinct accepted
edits are 51.6\% deletions of unused, no-op, or unreachable material,
47.6\% tactic replacements, and 0.8\% local fact generalizations.

\paragraph{Splits and leakage.}
Goedel-Workbook and Mathlib are training sources; miniF2F,
PutnamBench-verified, and Putnam~2025 are held out. Goal-hash overlap
between training and held-out sources is at most 0.15\% Jaccard, and
overlapping training rows are removed (App.~\ref{app:leakage}); deletion
rows carry no goal and are separated by source only.

\paragraph{Release format.}
Accepted edits and failed siblings are linked JSONL streams with a
Croissant manifest and datasheet; the repository adds a replay verifier
and all scripts behind the tables (schema in App.~\ref{app:schema};
errata in App.~\ref{app:errata}).

\paragraph{Symbolic headroom across proof sources.}
\label{sec:symbolic}
\label{sec:ablation-summary}
In its release run, the symbolic pass removes 19.7\% / 6.1\% of tokens
from Goedel-Prover-V2 proofs of miniF2F / PutnamBench-verified (27.5\% /
16.1\% when iterated to a fixed point), 5.5\% on
Goedel-Workbook, 1.3\% on AxiomProver, 0.27\% on Mathlib, and
2.6\% on Seed-Prover~1.5's Putnam~2025 release and 0.3--0.5\% on
Aristotle's IMO~2025 files, versus
${\le}0.01\%$ for Mathlib's \texttt{linter.unusedTactic} on identical
inputs (App.~\ref{app:symbolic-main}). Sources differ mainly in how
often they contain unused facts and speculative tactic cascades
(accepted edits per 1{,}000 tokens range $20\times$, savings per edit
$4\times$); tactic replacement and unused-fact removal account for most
savings, while local generalization is rare (0.8\% of edits) and has no demonstrated
aggregate compression benefit in these ablations.

\section{Selection Effects in Search-Generated Supervision}
\label{sec:selection}
\label{sec:learning}

A search that stops at its first success, and a benchmark built from the
sites where it succeeded, both encode the search's choices. We ask what
a model trained on these records learns, and which evaluations can tell.

\paragraph{Task.}
Given the goal state at a site, a window of surrounding proof text, and
the original span, a model must output a replacement span, or
\texttt{<DELETE>} (prompt in App.~\ref{app:learning}). Sites are those
where \LP{} (the \emph{teacher}) found a verified edit on the held-out sources; this is
\emph{conditional local editing} at teacher-identified sites, not
discovery of where to edit.

\paragraph{Data and models.}
From the 27{,}517 accepted edits of Goedel-Workbook and Mathlib we
remove 9{,}842 exact duplicates (same goal, original, and replacement),
27 edits that are not shorter in bytes, and 112 rows whose goal hash
occurs in a held-out source, and split the rest by source file into
17{,}099 training and 437 validation edits. We evaluate on all 1{,}406 held-out sites: 1{,}184 miniF2F, 80
PutnamBench-verified, and 142 Putnam~2025 (AxiomProver, a generator
absent from training). We fine-tune DeepSeek-Prover-V2-7B
\citep{ren2025deepseekv2} and Qwen2.5-Coder-7B-Instruct
\citep{hui2024qwen25coder}, a model not specialized for theorem
proving, with identical BF16 LoRA \citep{hu2022lora} settings (rank 32,
2 epochs, under one H100-hour). Three training seeds per model differ by
at most 1.4 points (standard deviation) on any held-out source.

\paragraph{Verification and metric.}
Each output is spliced into the original file at the site. A site
counts as a success (\emph{valid-and-shorter}) only if the spliced file
received a pass verdict from a fresh, byte-exact compilation under Lean
4.21.0 / Mathlib v4.21.0 and the replacement is strictly shorter in
tokens; reference-identical outputs are not assumed valid but are
re-verified. Failures save zero tokens. We report greedy decoding;
confidence intervals resample source files (10{,}000 replicates).
Token totals here sum individually verified local edits; whole-file
results in \S\ref{sec:use} instead verify and count the joint output.
Zero-shot frozen models use the same raw prompt and decoding; four-shot
controls additionally use each model's chat template.

\subsection{What fine-tuning learns at teacher-selected sites}

\begin{table}[t]
\caption{Greedy editing at teacher-selected sites. ``All'' and ``Tac.''
are valid-and-shorter rates (\%): 1{,}184 / 80 / 142 total sites and
747 / 43 / 57 tactic sites. ``Tok.'' sums individually verified savings;
failures save zero. The verified \emph{reference} can be byte-shorter
without being token-shorter. The \emph{delete rule} acts on the no-goal
marker. \emph{4-shot} uses the model's chat template and four fixed
examples (two replacements, two deletions). DPO starts from DeepSeek SFT.}
\label{tab:sft}
\centering
\small
\setlength{\tabcolsep}{4pt}
\resizebox{\linewidth}{!}{%
\begin{tabular}{lrrrcrrrcrrr}
\toprule
& \multicolumn{3}{c}{miniF2F} && \multicolumn{3}{c}{PutnamBench-verified} && \multicolumn{3}{c}{Putnam 2025 (AxiomProver)} \\
\cmidrule(lr){2-4}\cmidrule(lr){6-8}\cmidrule(lr){10-12}
Method & All & Tac. & Tok. && All & Tac. & Tok. && All & Tac. & Tok. \\
\midrule
Reference (pipeline) & 94.6 & 91.6 & 27{,}622 && 88.8 & 79.1 & 1{,}359 && 84.5 & 61.4 & 1{,}746 \\
Delete rule (no model) & 36.8 & 0.0 & 8{,}122 && 46.3 & 0.0 & 975 && 59.9 & 0.0 & 1{,}500 \\
\midrule
DeepSeek, frozen & 0.1 & 0.1 & 28 && 0.0 & 0.0 & 0 && 0.0 & 0.0 & 0 \\
Qwen, frozen & 0.5 & 0.8 & 189 && 0.0 & 0.0 & 0 && 1.4 & 0.0 & 32 \\
DeepSeek, 4-shot & 36.9 & 4.7 & 7{,}539 && 48.8 & 4.7 & 979 && 53.5 & 0.0 & 1{,}350 \\
Qwen, 4-shot & 46.9 & 17.9 & 9{,}046 && 48.8 & 7.0 & 964 && 59.2 & 12.3 & 1{,}401 \\
\midrule
DeepSeek, SFT & \textbf{86.7} & \textbf{79.0} & \textbf{26{,}163} && \textbf{86.3} & \textbf{74.4} & \textbf{1{,}337} && \textbf{79.6} & \textbf{49.1} & \textbf{1{,}666} \\
Qwen, SFT & 84.5 & 75.5 & 25{,}317 && 82.5 & 67.4 & 1{,}316 && 76.8 & 42.1 & 1{,}648 \\
DeepSeek, DPO & 84.5 & 75.6 & 25{,}614 && 85.0 & 72.1 & 1{,}333 && 73.2 & 33.3 & 1{,}618 \\
\bottomrule
\end{tabular}}
\end{table}

\paragraph{Fine-tuning improves conditional editing.}
Both frozen models almost never produce a valid shorter edit under
greedy decoding, while both fine-tuned models reach 77--87\% of all
sites and recover 92--98\% of the pipeline's savings on the
same sites (Table~\ref{tab:sft}; DeepSeek SFT file-bootstrap 95\% CIs
[83.4, 89.8] / [75.8, 96.6] / [58.2, 95.2]). The benefit is also present without prover-specific pretraining: Qwen2.5-Coder, a general code model, is
only 2.2 points lower on miniF2F (paired file bootstrap [0.8, 3.7]),
and differences on the two smaller sources are not resolved. The
near-zero zero-shot rates largely reflect the output interface. With
each model's chat template and four demonstrations, frozen models reach
37--59\% of all sites, but almost entirely through deletions: they
output \texttt{<DELETE>} at 72--78\% of sites and succeed at only
0--18\% of tactic-replacement sites, versus 42--79\% after fine-tuning.
These controls show that output formatting explains part of the
zero-shot gap, while fine-tuning substantially improves tactic replacement
over the tested prompts.

\paragraph{Separate deletions from tactic replacement.}
558 of the 1{,}406 sites are deletions of unused or unreachable
material, and at every one of them the prompt shows no local goal. A
rule that deletes exactly at those sites, with no model, matches all
558 references (each with a pass verdict) and alone yields 8{,}122 /
975 / 1{,}500 tokens: on Putnam~2025, 90\% of the fine-tuned model's
savings. Aggregate rates therefore mix a trivial decision with a harder one.
Tactic replacement measures performance beyond the deletion rule: the fine-tuned DeepSeek model
succeeds at 79.0\% / 74.4\% / 49.1\% of sites, against a reference that
is itself token-shorter at 91.6\% / 79.1\% / 61.4\%. Success is lower on AxiomProver proofs, whose generator and problem
set both differ from the training sources; this comparison alone does
not isolate the cause of the drop. The single
held-out generalization site is never solved and is too small to
evaluate learned abstraction.

\paragraph{Imitation, not improvement.}
Greedy exact match with the reference is 82.7\% / 82.5\% / 91.5\% for
DeepSeek SFT, and no fine-tuned output saves more tokens than its
reference. Three training seeds per model reproduce these rates
(standard deviation $\le$1.4 points). DPO on accepted/failed pairs \citep{rafailov2023dpo} trains
stably but is 2.1 points below SFT on miniF2F (paired $[1.0, 3.3]$). At
teacher-selected sites, the supervision teaches imitation of the
teacher.

\subsection{Ordering leakage, and complete pools that remove it}
\label{sec:ranking}

The failed siblings suggest a ranking task: given a goal and several
candidates, pick the one that works. A ranker trained on 54{,}375
goal-disjoint pairs reaches 100\% top-1 on all 718 held-out candidate
groups, far above random (21.8\%) and frozen log-probability (34.6\%).
This number is \emph{not} evidence of mathematical discrimination. By
construction (\S\ref{sec:method}) each group contains the winner and
the candidates tried \emph{before} it, so choosing the candidate latest
in the fixed menu order, using no goal, context, or model, also scores
100\% on all 718 groups, because the winner is always the
latest-tried candidate. The negatives thus encode the search policy, as in any first-success
search that logs its failures.

Preference pairs inherit the same shortcut. This limits what their
labels establish; our DPO comparison does not isolate it as the cause
of DPO's lower performance.

\begin{wraptable}{r}{0.44\linewidth}
\centering
\small
\setlength{\tabcolsep}{3pt}
\caption{Top-1 accuracy (\%) for the shortest valid, filter-passing candidate
on the evaluated complete pools (2{,}388 held-out states).
$^*$100\% on the biased first-success groups.}
\label{tab:pools}
\begin{tabular}{lr}
\toprule
Method & Top-1 \\
\midrule
Last in menu order$^*$ & 0.0 \\
First in menu order & 11.6 \\
Random & 8.2 \\
Shortest string & 5.8 \\
Per-tactic prior & 22.6 \\
Frozen Qwen-32B log-prob & 28.3 \\
Frozen DeepSeek-7B log-prob & 36.9 \\
Trained ranker, no goal & 51.0 \\
Trained ranker & \textbf{70.1} \\
\midrule
\emph{Ref.:} verifier, first success & 76.9 \\
\bottomrule
\end{tabular}
\end{wraptable}
\paragraph{Complete pools.}
Complete-menu search (\S\ref{sec:method}) removes this ordering
shortcut. The target is the shortest valid, filter-passing candidate
under the pool's string-length criterion. We built pools for 6{,}641 held-out states
(miniF2F, PutnamBench-verified, AxiomProver) and 53{,}272 states from
Goedel-Workbook and Mathlib, and trained a pointwise ranker on 13{,}290
Goedel states, with no goal hash shared with the held-out pools.
Table~\ref{tab:pools} evaluates the 2{,}388 held-out states whose pool
contains at least one acceptable candidate (valid, strictly shorter,
filter-passing); the other 4{,}253 states have no target and are excluded.
A choice counts as best if it ties the shortest acceptable candidate;
ties in a method's scores are broken uniformly at random (in expectation).
Re-running a subset at lower parallelism reproduced all 9{,}920 checked
validity labels, providing evidence of label stability for that subset.
On the evaluated pools
(Table~\ref{tab:pools}) the ordering rule that was perfect on the
biased groups selects the best candidate 0\% of the time, and
model-free rules stay below 23\%. Frozen models reach at most 36.9\%,
while the trained ranker reaches 70.1\% (file-clustered 95\% CI
66.3--73.9; validity AUROC 0.968) and recovers 79\% of the oracle's
character savings. Removing the goal from its input costs 19.1 points (CI
15.2--22.7), significantly on miniF2F and PutnamBench but not on
AxiomProver; the goal therefore supplies predictive information beyond the
remaining input. First-success verified search selects the pool's best
candidate more often (76.9\%). Exhaustive verification defines the
pool oracle; the learned ranker's accuracy does not establish a speed
advantage or remove the need to verify its choice.

\section{From Local Supervision to Whole-Proof Improvement}
\label{sec:use}
\label{sec:hybrid}

A local editor evaluated at teacher-selected sites inherits the teacher's
choice of where to act. We now remove that assistance: candidate sites
are extracted from complete proofs, and savings are measured on jointly
verified output files. We ask whether neural proposals add beyond the
symbolic search, and whether training is responsible for that gain.

\paragraph{Verified hybrid compression.}
Algorithm~\ref{alg:hybrid} runs \LP{}, then asks a local editor (a trained or frozen 7B
model, prompted without an explicit symbolic menu) for
proposals at every remaining tactic site, verifies each in a fresh Lean
process, combines compatible edits, and checks the resulting file again.
This final check matters: edits that work separately may interfere when
applied together.

\subsection{Symbolic fixed point versus neural edits}
\label{sec:hybrid-results}

\begin{table}[t]
\caption{Whole-file token reduction (\%) of Lean-verified final files, on
specified input sets (Lean-aware counter; unshortened files count 0).
miniF2F-99 is a fixed random 100-file subset of the 351 files, minus one
linter-baseline artifact (App.~\ref{app:errata}). \emph{Release
run}: the \LP{} run that produced the dataset; \emph{clean rerun}: the
same binary re-run without resource contention. Hybrid rows: best of greedy and four samples; frozen models are
DeepSeek-Prover-V2-7B unless stated.
\emph{filter}: only edits that pass \LP{}'s specificity filter. LLM rewrites use $k{=}16$ samples.}
\label{tab:hybrid}
\centering
\small
\setlength{\tabcolsep}{4pt}
\begin{tabular}{lrrr}
\toprule
Method & AxiomProver (12) & PutnamBench (19) & miniF2F-99 \\
\midrule
\multicolumn{4}{l}{\emph{Symbolic only}} \\
\quad \LP{}, release run & 1.31 & 6.14 & 20.02 \\
\quad \LP{}, clean rerun & 1.41 & 11.91 & 20.52 \\
\quad \LP{} (clean) iterated to a fixed point & 1.76 & 16.07 & \textbf{29.76} \\
\multicolumn{4}{l}{\emph{Neural only}} \\
\quad LLM whole-proof rewrite, frozen & 1.04$^a$ & 2.76 & 12.52 \\
\quad LLM rewrite, trained on \LP{} pairs & ---$^d$ & 5.53 & 16.36 \\
\quad Local editor alone (trained) & 2.51 & 3.10 & --- \\
\multicolumn{4}{l}{\emph{Hybrid (Alg.~\ref{alg:hybrid}): release run, then verified local editor}} \\
\quad frozen, few-shot & 3.66 & 9.31 & 23.14 \\
\quad frozen, few-shot, filter & 3.66 & 9.30 & 23.13 \\
\quad trained DeepSeek & 3.65 & 9.02 & 22.41 \\
\quad trained DeepSeek, filter & 2.38 & 8.04 & 21.37 \\
\quad trained Qwen & 4.00 & 11.38 & ---$^c$ \\
\multicolumn{4}{l}{\emph{Composed pipelines}} \\
\quad \LP{} (release) $\to$ frozen LLM rewrite & 2.34$^a$ & 7.88 & 27.58 \\
\quad \LP{} (release) $\to$ trained LLM rewrite & ---$^d$ & 8.36 & 27.13 \\
\quad \LP{} (clean) $\to$ frozen LLM rewrite & --- & 12.85 & 27.11 \\
\quad frozen LLM rewrite $\to$ \LP{} & 2.53$^a$ & 10.36 & 27.99 \\
\quad \LP{} $\leftrightarrow$ trained editor, alternated & \textbf{5.05}$^b$ & \textbf{20.25} & --- \\
\bottomrule
\end{tabular}
\par\smallskip{\raggedright\footnotesize $^a$$k{=}4$. $^b$Two rounds (PutnamBench: three). $^c$23.76 on all 351 miniF2F files (DeepSeek: 21.74). $^d$Only 3 of 12 AxiomProver files fit the rewriter's context; no gain on them.\par}
\end{table}

\paragraph{Iterate the symbolic baseline before attributing neural gains.}
Table~\ref{tab:hybrid} compares complete output files rather than sums of
local edits. On miniF2F-99, a single symbolic pass removes 20.0--20.5\%
of tokens, versus 12.5\% for frozen whole-proof rewriting. Adding a local
editor or whole-proof rewriter improves on that single pass. However,
symbolic search also improves when given another opportunity: a clean
rerun raises PutnamBench reduction from 6.1\% to 11.9\%, and iteration
to a fixed point reaches 16.1\% on PutnamBench and 29.8\% on miniF2F-99
(27.5\% on all 351 miniF2F files). The miniF2F fixed point exceeds every
tested neural hybrid on the corresponding input set. A one-pass teacher
therefore understates the available symbolic baseline.

Neural editing still adds on other sources. Alternating \LP{} and the
trained editor reaches 20.3\% on PutnamBench, 4.2 percentage points above
the symbolic fixed point, and 5.05\% on AxiomProver, versus 1.76\%
symbolically. A single hybrid pass also improves six Seed-Prover~1.5
files from 4.37\% to 6.36\%. These establish additional compression in
the tested configurations, not a compute-normalized advantage: the
pipelines differ in model calls, verification work, and iteration count.

\subsection{What training contributes}
\label{sec:training-utility}

\paragraph{Local editing: a strong frozen control.}
With the same sampling and verification loop, a frozen DeepSeek 7B
editor with four demonstrations reaches 9.31\% on PutnamBench,
23.14\% on miniF2F-99, and 3.66\% on AxiomProver, compared with
9.02\%, 22.41\%, and 3.65\% for the trained editor. Thus these
whole-file comparisons do not establish an advantage from fine-tuning,
despite its large benefit at teacher-selected sites. The editors also
act differently: 397 of the frozen editor's 414 applied AxiomProver
edits delete redundant tactic lines, whereas the trained editor makes
substitutions and no deletions.

Compression also depends on the edit policy. About 97\% of the trained
editor's accepted edits use tactics already in the symbolic menu; on
AxiomProver, 526 of its 753 in-menu edits violate \LP{}'s specificity
filter, commonly replacing an explicit term by automation. Restricting
the DeepSeek hybrid to filter-compliant edits reduces its AxiomProver
savings from 3.65\% to 2.38\%, still above the 1.31\% release baseline.
The frozen editor's deletions pass the same filter. These controls
separate gains from broader proposals from gains obtained by relaxing
the teacher's policy. Qwen2.5-Coder trained on the same records yields
higher hybrid reductions than DeepSeek on all three evaluated sources,
so prover-specific pretraining is not necessary for this use.

\paragraph{Whole-proof rewriting: a benefit from the supervision.}
Fine-tuning DeepSeek-Prover-V2-7B for 67 GPU-minutes on 7{,}708
Goedel-Workbook proof pairs (original $\to$ \LP{} output, 11\%
unchanged) raises verified token reduction at $k{=}16$ from 2.8\% to
5.5\% on the same 19 PutnamBench files. On miniF2F, the reported
reductions on the same 99 files are 12.5\% for the frozen rewriter and
16.4\% after training, and 62\% of the trained model's samples yield a
verified shorter file, against 16\% for the frozen model
(Table~\ref{tab:hybrid}); theorem statements are unchanged. The trained
rewriter remains below the symbolic fixed point,
and training adds no consistent gain when rewriting follows \LP{}
(8.36\% vs.\ 7.88\% on PutnamBench; 27.13\% vs.\ 27.58\% on
miniF2F-99). This supports transfer of the teacher's edits to
a whole-proof interface, without establishing improvement beyond the
teacher's remaining search space.

\paragraph{One round of self-training.}
\label{sec:selfimprove}
We also test whether the editor's own verified outputs improve it further
\citep{polu2023curriculum, ahuja2026improver2}. From 5{,}472 unchanged
tactic sites in 1{,}000 training proofs, we obtain 372 distinct new
edits; 147 violate the specificity filter. Continued training on either
all new edits or filter-compliant non-deletions, mixed with replay,
reduces success on the 847 held-out tactic sites from 74.9\% to 72.9\%
or 70.7\%, respectively (paired 95\% CIs for the changes:
$[-3.6,-0.5]$ and $[-6.1,-2.2]$ points). Whole-file savings beyond the
teacher do not change significantly on PutnamBench or AxiomProver.
This small, one-round experiment finds no benefit from further
self-training; it does not establish a general limit on expert iteration
(details in App.~\ref{app:learning}).


\section{Discussion}
\label{sec:discussion}
\label{sec:limitations}

\LP{} makes proof-improvement supervision inspectable at the level where
an edit is proposed and verified. The resulting resource exposes a
distinction that aggregate accuracy obscures: a model can reproduce a
search procedure's choices without learning to improve on that search.
Complete-menu outcomes remove one concrete shortcut, and the goal
ablation shows useful state-dependent information in the remaining
task. Whole-proof fine-tuning on PutnamBench provides a complementary
positive result: symbolic edits can train a model to produce more
effective verified rewrites.

The broader lesson is to evaluate the source of an improvement. A
symbolic fixed point controls for edits missed by an early stop; a
frozen editor controls for generate-and-verify alone; and a shared
specificity policy distinguishes broader search from a different
definition of an acceptable edit. These controls materially change our
conclusions. They also give future work a concrete target: improve
complete proofs beyond strong symbolic and prompted baselines at a
declared computational budget.

The evidence has limits. Whole-file comparisons use 12--351 files per
source and one trained checkpoint per configuration; three-seed checks
cover conditional local editing only. Search remains bounded by its
menu, imports, and time budgets, and complete pools do not remove all
dataset selection effects. We measure length and compliance with a
specificity policy, not human readability, maintenance cost, or improved
theorem-proving ability. Within this scope, \LP{} contributes a verified neurosymbolic compression method, its
supervision, and a protocol for testing what that supervision teaches. Verification certifies correctness; controlled comparisons show when
learning helps.


\subsubsection*{Code and Data}
The code repository
(\url{https://github.com/paulinebourigault/leanpolish}) contains the pipeline (including the
complete-menu patch), the verifier, both token counters, and every
script behind the tables, with a README mapping each table to its
command and result files; all reported numbers are recomputed by these
scripts from saved outputs over exact input-file lists. The dataset, the
complete candidate pools, all model generations with their Lean
verdicts, the verified output files, and the LoRA adapters of the
editors are at \url{\dataurl} (folders \texttt{experiments/} and
\texttt{adapters/}). Toolchains are pinned (Lean~4.21.0, Mathlib
v4.21.0); training hyperparameters and prompts are in
App.~\ref{app:learning}. Symbolic results can vary with machine speed
through per-candidate time budgets (App.~\ref{app:errata}); we report
both the release run and a clean rerun.

%

\bibliography{references}

@inproceedings{gu2025proofoptimizer,
  title     = {ProofOptimizer: Training Language Models to Simplify Proofs without Human Demonstrations},
  author    = {Gu, Alex and Piotrowski, Bartosz and Gloeckle, Fabian and Yang, Kaiyu and Markosyan, Aram H.},
  booktitle = {International Conference on Learning Representations},
  year      = {2026}
}

@inproceedings{yang2023leandojo,
  title     = {LeanDojo: Theorem Proving with Retrieval-Augmented Language Models},
  author    = {Yang, Kaiyu and Swope, Aidan and Gu, Alex and Chalamala, Rahul and Song, Peiyang and Yu, Shixing and Godil, Saad and Prenger, Ryan J. and Anandkumar, Anima},
  booktitle = {Advances in Neural Information Processing Systems (NeurIPS)},
  year      = {2023},
}

@misc{lin2025goedelv1,
  title         = {Goedel-Prover: A Frontier Model for Open-Source Automated Theorem Proving},
  author        = {Lin, Yong and Tang, Shange and Lyu, Bohan and Wu, Jiayun and Lin, Hongzhou and Yang, Kaiyu and Li, Jia and Xia, Mengzhou and Chen, Danqi and Arora, Sanjeev and Jin, Chi},
  year          = {2025},
  eprint        = {2502.07640},
  archivePrefix = {arXiv},
  primaryClass  = {cs.LG},
}

@misc{lin2025goedelv2,
  title         = {Goedel-Prover-V2: Scaling Formal Theorem Proving with Scaffolded Data Synthesis and Self-Correction},
  author        = {Lin, Yong and Tang, Shange and Lyu, Bohan and Yang, Ziran and Chung, Jui-Hui and Zhao, Haoyu and Jiang, Lai and Geng, Yihan and Ge, Jiawei and Sun, Jingruo and Wu, Jiayun and Gesi, Jiri and Lu, Ximing and Acuna, David and Yang, Kaiyu and Lin, Hongzhou and Choi, Yejin and Chen, Danqi and Arora, Sanjeev and Jin, Chi},
  year          = {2025},
  eprint        = {2508.03613},
  archivePrefix = {arXiv},
  primaryClass  = {cs.LG},
}

@inproceedings{lample2022htps,
  title     = {{HyperTree} Proof Search for Neural Theorem Proving},
  author    = {Lample, Guillaume and Lacroix, Timoth{\'e}e and Lachaux, Marie-Anne and Rodr{\'\i}guez, Aur{\'e}lien and Hayat, Amaury and Lavril, Thibaut and Ebner, Gabriel and Martinet, Xavier},
  booktitle = {Advances in Neural Information Processing Systems (NeurIPS)},
  year      = {2022},
}

@misc{ren2025deepseekv2,
  title         = {{DeepSeek-Prover-V2}: Advancing Formal Mathematical Reasoning via Reinforcement Learning for Subgoal Decomposition},
  author        = {Ren, Z. Z. and Shao, Zhihong and Song, Junxiao and Xin, Huajian and Wang, Haocheng and Zhao, Wanjia and Zhang, Liyue and Fu, Zhe and Zhu, Qihao and Yang, Dejian and Wu, Z.~F. and Gou, Zhibin and Ma, Shirong and Tang, Hongxuan and Liu, Yuxuan and Gao, Wenjun and Guo, Daya and Ruan, Chong},
  year          = {2025},
  eprint        = {2504.21801},
  archivePrefix = {arXiv},
  primaryClass  = {cs.LG},
}

@misc{achim2025aristotle,
  title         = {{Aristotle}: {IMO}-level Automated Theorem Proving},
  author        = {Achim, Tudor and Best, Alex and Der, Kevin and F{\'e}d{\'e}rico, Math{\"i}s and Gukov, Sergei and Halpern-Leister, Daniel and Henningsgard, Kirsten and Kudryashov, Yury and Meiburg, Alexander and Michelsen, Martin and Patterson, Riley and Rodriguez, Eric and Scharff, Laura and Shanker, Vikram and Sicca, Vladmir and Sowrirajan, Hari and Swope, Aidan and Tamas, Matyas and Tenev, Vlad and Thomm, Jonathan and Williams, Harold and Wu, Lawrence},
  year          = {2025},
  eprint        = {2510.01346},
  archivePrefix = {arXiv},
  primaryClass  = {cs.LG},
  note          = {The Harmonic Team.},
}

@misc{wang2025kimina,
  title         = {{Kimina-Prover} Preview: Towards Large Formal Reasoning Models with Reinforcement Learning},
  author        = {Wang, Haiming and Unsal, Mert and Lin, Xiaohan and Baksys, Mantas and Liu, Junqi and Santos, Marco Dos and Sung, Flood and Vinyes, Marina and Ying, Zhenzhen and Zhu, Zekai and Lu, Jianqiao and Saxcé, Hugues de and Bailey, Bolton and Song, Chendong and Xiao, Chenjun and Zhang, Dehao and Zhang, Ebony and Pu, Frederick and Zhu, Han and Liu, Jiawei and Bayer, Jonas and Michaud, Julien and Hu, Kaiyu and Pernigo, Kashun and Liang, Kjartan Kaarstan and Wang, Lingxiao and Wu, Mingfei and Hu, Tao and Yan, Tianyu and Tsai, Yifei and Aliyev, Abulay and Kushnirskyi, Bohdan and Bohush, Igor and Yan, Boyang and Zhu, Yifan and Yu, Yujian and Pan, Junyu and Yang, Yongqian and Zhang, Yuhua and Liu, Zhengying and Li, Jia},
  year          = {2025},
  eprint        = {2504.11354},
  archivePrefix = {arXiv},
  primaryClass  = {cs.AI},
}

@misc{axiom2025putnam,
  author       = {{Axiom Math}},
  title        = {{AxiomProver}: An Autonomous Multi-Agent Ensemble Theorem Prover for {Lean 4}: {Putnam} 2025 Solutions},
  year         = {2025},
  howpublished = {GitHub repository},
  note         = {12/12 problems on the {Putnam} 2025 competition. \url{https://github.com/AxiomMath/Putnam2025}},
}

@inproceedings{ying2024leanworkbook,
  title     = {Lean Workbook: A Large-Scale {Lean} Problem Set Formalized from Natural Language Math Problems},
  author    = {Ying, Huaiyuan and Wu, Zijian and Geng, Yihan and Wang, Jiayu and Lin, Dahua and Chen, Kai},
  booktitle = {Advances in Neural Information Processing Systems (NeurIPS)},
  year      = {2024},
  pages     = {105848--105863},
}

@incollection{plotkin1970,
  author    = {Plotkin, G. D.},
  title     = {A Note on Inductive Generalization},
  booktitle = {Machine Intelligence 5},
  editor    = {Meltzer, B. and Michie, D.},
  publisher = {Edinburgh University Press},
  year      = {1970},
  pages     = {153--163},
}

@article{cerna2023antiunification,
  author  = {Cerna, David M. and Kutsia, Temur},
  title   = {Anti-Unification and Generalization: A Survey},
  journal = {Journal of Artificial Intelligence Research},
  volume  = {78},
  year    = {2023},
  pages   = {293--361},
}

@inproceedings{rafailov2023dpo,
  title     = {Direct Preference Optimization: Your Language Model is Secretly a Reward Model},
  author    = {Rafailov, Rafael and Sharma, Archit and Mitchell, Eric and Manning, Christopher D. and Ermon, Stefano and Finn, Chelsea},
  booktitle = {Advances in Neural Information Processing Systems (NeurIPS)},
  year      = {2023},
}

@inproceedings{zheng2022minif2f,
  title     = {{MiniF2F}: A Cross-System Benchmark for Formal Olympiad-Level Mathematics},
  author    = {Zheng, Kunhao and Han, Jesse Michael and Polu, Stanislas},
  booktitle = {International Conference on Learning Representations (ICLR)},
  year      = {2022},
}

@inproceedings{tsoukalas2024putnambench,
  title     = {{PutnamBench}: Evaluating Neural Theorem-Provers on the {Putnam} Mathematical Competition},
  author    = {Tsoukalas, George and Lee, Jasper and Jennings, John and Xin, Jimmy and Ding, Michelle and Jennings, Michael and Thakur, Amitayush and Chaudhuri, Swarat},
  booktitle = {Advances in Neural Information Processing Systems (NeurIPS) Datasets and Benchmarks Track},
  year      = {2024},
}

@inproceedings{demoura2021lean4,
  title     = {The {Lean 4} Theorem Prover and Programming Language},
  author    = {de Moura, Leonardo and Ullrich, Sebastian},
  booktitle = {Automated Deduction -- {CADE} 28},
  year      = {2021},
  pages     = {625--635},
}

@misc{aristotle2025imo,
  title        = {Aristotle: Official Lean 4 Solutions to {IMO} 2025},
  author       = {{Harmonic}},
  year         = {2025},
  note         = {Official formal-solution release}
}

@misc{hui2024qwen25coder,
  title         = {Qwen2.5-Coder Technical Report},
  author        = {Binyuan Hui and Jian Yang and Zeyu Cui and Jiaxi Yang and Dayiheng Liu and Lei Zhang and Tianyu Liu and Jiajun Zhang and Bowen Yu and Keming Lu and others},
  year          = {2024},
  eprint        = {2409.12186},
  archivePrefix = {arXiv}
}

@inproceedings{hu2022lora,
  title     = {{LoRA}: Low-Rank Adaptation of Large Language Models},
  author    = {Edward J. Hu and Yelong Shen and Phillip Wallis and Zeyuan Allen-Zhu and Yuanzhi Li and Shean Wang and Lu Wang and Weizhu Chen},
  booktitle = {International Conference on Learning Representations},
  year      = {2022}
}

@inproceedings{ahuja2025improver,
  title     = {{ImProver}: Agent-Based Automated Proof Optimization},
  author    = {Riyaz Ahuja and Jeremy Avigad and Prasad Tetali and Sean Welleck},
  booktitle = {International Conference on Learning Representations},
  year      = {2025}
}

@misc{ahuja2026improver2,
  title         = {{ImProver} 2: Iteratively Self-Improving {LMs} for Neurosymbolic Proof Optimization},
  author        = {Ahuja, Riyaz and Rowney, Tate and Avigad, Jeremy and Welleck, Sean},
  year          = {2026},
  eprint        = {2605.22885},
  archivePrefix = {arXiv},
  primaryClass  = {cs.AI},
}

@misc{lu2026leanrefactor,
  title         = {Lean Refactor: Multi-Objective Controllable Proof Optimization via Agentic Strategy Search},
  author        = {Lu, Jialin and Kong, Soonho and Stehling, Rodrigo and Yang, Kaiyu and Wang, Zhangyang and Sun, Weiran and Chen, Wuyang},
  year          = {2026},
  eprint        = {2605.20244},
  archivePrefix = {arXiv},
  primaryClass  = {cs.LO},
}

@misc{fu2026proofrefactor,
  title         = {Proof-Refactor: Refactoring Generated Formal Proofs into Modular Artifacts},
  author        = {Fu, Yiming and Liu, Peixuan and Wang, Zichen and Yuan, Kun},
  year          = {2026},
  eprint        = {2606.03743},
  archivePrefix = {arXiv},
  primaryClass  = {cs.AI},
}

@misc{xin2026axle,
  title         = {{AXLE}: A Cloud Infrastructure for {Lean 4} Theorem Proving Utilities},
  author        = {Xin, Jimmy and Schneidman, Alex and Cummins, Chris and Ram, Karun and Ganesh, Srihari and Limperg, Jannis},
  year          = {2026},
  eprint        = {2606.26442},
  archivePrefix = {arXiv},
}

@misc{wang2026april,
  title         = {Learning to Repair {Lean} Proofs from Compiler Feedback},
  author        = {Wang, Evan and Chess, Simon and Lee, Daniel and Ge, Siyuan and Mallavarapu, Ajit and Alper, Jarod and Ilin, Vasily},
  year          = {2026},
  eprint        = {2602.02990},
  archivePrefix = {arXiv},
  note          = {ICLR 2026 VerifAI Workshop},
}

@inproceedings{ospanov2025apollo,
  title     = {{APOLLO}: Automated {LLM} and {Lean} Collaboration for Advanced Formal Reasoning},
  author    = {Ospanov, Azim and Farnia, Farzan and Yousefzadeh, Roozbeh},
  booktitle = {Advances in Neural Information Processing Systems (NeurIPS)},
  year      = {2025},
  note      = {arXiv:2505.05758},
}

@misc{khrulev2026blueprintrepair,
  title         = {{BlueprintRepair}: Typed Local Edits for Failed {Lean} Proof Blueprints},
  author        = {Khrulev, Ruslan},
  year          = {2026},
  eprint        = {2607.28110},
  archivePrefix = {arXiv},
  primaryClass  = {cs.AI},
}

@inproceedings{first2023baldur,
  title     = {Baldur: Whole-Proof Generation and Repair with Large Language Models},
  author    = {First, Emily and Rabe, Markus N. and Ringer, Talia and Brun, Yuriy},
  booktitle = {Proceedings of the 31st ACM Joint European Software Engineering Conference and Symposium on the Foundations of Software Engineering (ESEC/FSE)},
  year      = {2023},
  doi       = {10.1145/3611643.3616243},
}

@misc{liu2025proofaug,
  title         = {{ProofAug}: Efficient Neural Theorem Proving via Fine-grained Proof Structure Analysis},
  author        = {Liu, Haoxiong and Sun, Jiacheng and Li, Zhenguo and Yao, Andrew C.},
  year          = {2025},
  eprint        = {2501.18310},
  archivePrefix = {arXiv},
}

@inproceedings{polu2023curriculum,
  title     = {Formal Mathematics Statement Curriculum Learning},
  author    = {Polu, Stanislas and Han, Jesse Michael and Zheng, Kunhao and Baksys, Mantas and Babuschkin, Igor and Sutskever, Ilya},
  booktitle = {International Conference on Learning Representations (ICLR)},
  year      = {2023},
  note      = {arXiv:2202.01344},
}

@article{hubert2025alphaproof,
  title   = {Olympiad-level formal mathematical reasoning with reinforcement learning},
  author  = {Hubert, Thomas and Mehta, Rishi and Sartran, Laurent and Horv{\'a}th, Mikl{\'o}s Z. and {\v{Z}}u{\v{z}}i{\'c}, Goran and Wieser, Eric and Huang, Aja and Schrittwieser, Julian and others},
  journal = {Nature},
  volume  = {651},
  pages   = {607--613},
  year    = {2025},
  doi     = {10.1038/s41586-025-09833-y},
}

@misc{xin2024deepseekv15,
  title         = {{DeepSeek-Prover-V1.5}: Harnessing Proof Assistant Feedback for Reinforcement Learning and {Monte-Carlo} Tree Search},
  author        = {Xin, Huajian and Ren, Z. Z. and Song, Junxiao and Shao, Zhihong and Zhao, Wanjia and Wang, Haocheng and Liu, Bo and Zhang, Liyue and Lu, Xuan and Du, Qiushi and Gao, Wenjun and Zhu, Qihao and Yang, Dejian and Gou, Zhibin and Wu, Z. F. and Luo, Fuli and Ruan, Chong},
  year          = {2024},
  eprint        = {2408.08152},
  archivePrefix = {arXiv},
}

@inproceedings{xin2025bfsprover,
  title     = {{BFS-Prover}: Scalable Best-First Tree Search for {LLM}-based Automatic Theorem Proving},
  author    = {Xin, Ran and Xi, Chenguang and Yang, Jie and Chen, Feng and Wu, Hang and Xiao, Xia and Sun, Yifan and Zheng, Shen and Shen, Kai},
  booktitle = {Proceedings of the 63rd Annual Meeting of the Association for Computational Linguistics (ACL)},
  year      = {2025},
  note      = {arXiv:2502.03438},
}

@misc{chen2025seedprover15,
  title         = {{Seed-Prover} 1.5: Mastering Undergraduate-Level Theorem Proving via Learning from Experience},
  author        = {Chen, Jiangjie and Chen, Wenxiang and Du, Jiacheng and Hu, Jinyi and Jiang, Zhicheng and others},
  year          = {2025},
  eprint        = {2512.17260},
  archivePrefix = {arXiv},
}

@inproceedings{an2024learnfromfailure,
  title     = {Learn from Failure: Fine-Tuning {LLMs} with Trial-and-Error Data for Intuitionistic Propositional Logic Proving},
  author    = {An, Chenyang and Chen, Zhibo and Ye, Qihao and First, Emily and Peng, Letian and Zhang, Jiayun and Wang, Zihan and Lerner, Sorin and Shang, Jingbo},
  booktitle = {Proceedings of the 62nd Annual Meeting of the Association for Computational Linguistics (ACL)},
  year      = {2024},
  note      = {arXiv:2404.07382},
}

@misc{kim2026processverified,
  title         = {Process-Verified Reinforcement Learning for Theorem Proving via {Lean}},
  author        = {Kim, Minsu and Yun, Se-Young},
  year          = {2026},
  eprint        = {2606.20068},
  archivePrefix = {arXiv},
}

@inproceedings{gao2025herald,
  title     = {Herald: A Natural Language Annotated {Lean 4} Dataset},
  author    = {Gao, Guoxiong and Wang, Yutong and Jiang, Jiedong and Gao, Qi and Qin, Zihan and Xu, Tianyi and Dong, Bin},
  booktitle = {International Conference on Learning Representations (ICLR)},
  year      = {2025},
  note      = {arXiv:2410.10878},
}

@misc{yu2025formalmath,
  title         = {{FormalMATH}: Benchmarking Formal Mathematical Reasoning of Large Language Models},
  author        = {Yu, Zhouliang and Peng, Ruotian and Ding, Keyi and Li, Yizhe and Peng, Zhongyuan and Liu, Minghao and Zhang, Yifan and Yuan, Zheng and Xin, Huajian and Huang, Wenhao and Wen, Yandong and Zhang, Ge and Liu, Weiyang},
  year          = {2025},
  eprint        = {2505.02735},
  archivePrefix = {arXiv},
}

@inproceedings{shypula2024pie,
  title     = {Learning Performance-Improving Code Edits},
  author    = {Shypula, Alexander and Madaan, Aman and Zeng, Yimeng and Alon, Uri and Gardner, Jacob and Hashemi, Milad and Neubig, Graham and Ranganathan, Parthasarathy and Bastani, Osbert and Yazdanbakhsh, Amir},
  booktitle = {International Conference on Learning Representations (ICLR)},
  year      = {2024},
  note      = {arXiv:2302.07867},
}

@inproceedings{schkufza2013stoke,
  title     = {Stochastic Superoptimization},
  author    = {Schkufza, Eric and Sharma, Rahul and Aiken, Alex},
  booktitle = {Proceedings of the 18th International Conference on Architectural Support for Programming Languages and Operating Systems (ASPLOS)},
  pages     = {305--316},
  year      = {2013},
  doi       = {10.1145/2451116.2451150},
}

@inproceedings{bunel2017superopt,
  title     = {Learning to Superoptimize Programs},
  author    = {Bunel, Rudy and Desmaison, Alban and Kumar, M. Pawan and Torr, Philip H. S. and Kohli, Pushmeet},
  booktitle = {International Conference on Learning Representations (ICLR)},
  year      = {2017},
  note      = {arXiv:1611.01787},
}

@misc{cummins2023llmopt,
  title         = {Large Language Models for Compiler Optimization},
  author        = {Cummins, Chris and Seeker, Volker and Grubisic, Dejan and Elhoushi, Mostafa and Liang, Youwei and Roziere, Baptiste and Gehring, Jonas and Gloeckle, Fabian and Hazelwood, Kim and Synnaeve, Gabriel and Leather, Hugh},
  year          = {2023},
  eprint        = {2309.07062},
  archivePrefix = {arXiv},
}

@inproceedings{cummins2025llmcompiler,
  title     = {{LLM} Compiler: Foundation Language Models for Compiler Optimization},
  author    = {Cummins, Chris and Seeker, Volker and Grubisic, Dejan and Roziere, Baptiste and Gehring, Jonas and Synnaeve, Gabriel and Leather, Hugh},
  booktitle = {Proceedings of the 34th ACM SIGPLAN International Conference on Compiler Construction (CC)},
  year      = {2025},
  doi       = {10.1145/3708493.3712691},
  note      = {arXiv:2407.02524},
}
\bibliographystyle{iclr2027_conference}

\newpage
\appendix
\section*{Appendix}
\startcontents[appendix]
\printcontents[appendix]{}{1}{\setcounter{tocdepth}{1}}
\bigskip

\section{Symbolic Compression Across Proof Sources}
\label{app:symbolic-main}

This section supports the cross-source results in \S\ref{sec:resource}:
it defines the compression metric, distinguishes matched baselines from
published contextual comparisons, and interprets the phase ablations.

\paragraph{Metric.}
We count tokens with a Lean-aware counter following the rules of
Appendix~L of \citet{gu2025proofoptimizer}: identifiers, literals, and
multi-character operators count as one token, and whitespace and
comments are skipped. This is the counter \LP{} uses internally; our
released Python port reproduces its per-file counts exactly. For an input set $\mathcal F$, whole-corpus token reduction is
$100\,\sum_{f\in\mathcal F}[L(f)-L(\hat f)]/\sum_{f\in\mathcal F}L(f)$,
where $L$ is the token count and $\hat f=f$ when no shortened output is
accepted. This weights files by original length; it is not the mean of
per-proof percentage reductions. Exact input-file lists are released. A second
released tokenizer that also counts comments gives similar numbers
except on comment-heavy sources (Goedel-Workbook: 9.5\% instead of
5.5\%, because removed proof text often carries comments); both counters are released, and results using the second counter are
identified explicitly. We also report bytes and non-blank lines
(App.~\ref{app:symbolic}); they agree on the extremes but rank
Goedel-Workbook higher, because removed text there often carries comments.

\paragraph{Matched-input comparison with the linter.}
Mathlib's \texttt{linter.unusedTactic} flags top-level tactics whose
removal leaves the proof state unchanged; it is the symbolic first
stage of ProofOptimizer, and the only published component we can run
on identical inputs. We reproduce its configuration and apply both
tools to the same files, toolchain, and tokenizer
(Table~\ref{tab:linter}). On Goedel-Prover-V2 miniF2F and PutnamBench
proofs, \LP{} removes 19.7\% and 6.1\% of tokens versus ${\le}0.002\%$
for the linter; on Mathlib, where the linter already runs in CI, both
are near zero. ProofOptimizer reports 9.2\% / 7.4\% for the linter on
194 / 75 Goedel-Prover-V2 proofs under Mathlib 4.19; under our v4.21
configuration it rarely fires on our 351 / 19 proofs. We have not
isolated whether the proof samples or the Mathlib version account for
this difference.

\begin{table}[t]
\caption{Whole-corpus token reduction (\%) on identical inputs,
toolchain, and tokenizer. ProofOptimizer's figures are published
per-proof average reductions on its own Goedel-Prover-V2 inputs for the
same benchmarks, shown for reference only (not a head-to-head
comparison).}
\label{tab:linter}
\centering
\small
\setlength{\tabcolsep}{4pt}
\begin{tabular}{lrrr}
\toprule
Input proofs (files shortened / total) & \LP{} & \texttt{unusedTactic} linter & ProofOptimizer (ref.) \\
\midrule
miniF2F, Goedel-Prover-V2 (308/351) & 19.7 & 0.002 & 87.9 \\
PutnamBench, Goedel-Prover-V2 (16/19) & 6.1 & 0.000 & 57.2 \\
Goedel-Workbook & 5.48 & 0.010 & --- \\
Mathlib v4.21.0 subset & 0.27 & 0.000 & --- \\
Putnam 2025, AxiomProver & 1.31 & 0.000 & --- \\
\bottomrule
\end{tabular}
\end{table}

\paragraph{Frontier-prover releases.}
To test whether symbolic headroom persists on proofs from newer
agentic systems, we ran the unchanged pipeline on the official
Seed-Prover~1.5 Putnam~2025 solutions \citep{chen2025seedprover15} and
Aristotle's IMO~2025 solutions \citep{aristotle2025imo}, without
modifying any input. Seed-Prover pins Lean/Mathlib 4.22, so for a
matched comparison we ran both Seed-Prover and AxiomProver under a
common 4.22 toolchain on the eight problems where both releases
compile; the port changes one heartbeat option in the optimizer and the
per-file time budget, not the inputs.
Seed-Prover proofs shrink by 2.7\% (3{,}803 / 139{,}490 tokens) versus
1.2\% (966 / 80{,}107) for AxiomProver, with per-problem reductions
from 0\% to 6.2\% (App.~\ref{app:frontier}). Under each release's
native toolchain, all eleven Seed-Prover files give 2.55\%, and the
full AxiomProver release gives 1.31\% (Table~\ref{tab:corpora}). Aristotle's two processable files
shrink by 0.3\% and 0.5\%; one file is toolchain-incompatible and one
103\,KB file triggers a deterministic indexing failure in the optimizer,
which we report as a robustness limitation. These are small samples
and independent formalizations of the same problems, so we claim no
trend; they show that headroom on frontier outputs is real but modest.

\begin{table}[t]
\caption{Descriptive edit statistics for the release run in
Table~\ref{tab:corpora}. Density is accepted edits per 1,000 original
tokens; tok/edit is mean tokens saved per accepted edit. Putnam 2025
uses the sequential configuration. These summarize the edit mix, not a
causal model of compression.}
\label{tab:edit-density}
\centering
\small
\begin{tabular}{lrr}
\toprule
Source & Density & Tok/edit \\
\midrule
Mathlib v4.21.0 subset & 0.46 & 5.9 \\
Goedel-Workbook & 4.72 & 11.6 \\
PutnamBench sample & 9.10 & 8.9 \\
miniF2F verified & 8.45 & 23.3 \\
PutnamBench verified & 3.62 & 17.0 \\
Putnam 2025 (AxiomProver) & 1.06 & 12.3 \\
\bottomrule
\end{tabular}
\end{table}

\paragraph{Why reductions vary.}
For non-overlapping edits whose savings sum to the file-level change,
accepted-edit density (per 1{,}000 original tokens) times mean tokens
saved per edit, divided by 1{,}000, gives the reduction as a fraction.
These descriptors help interpret the file-level metric; overlapping
local edits must not be summed as a whole-file result. Density spans
$20\times$ across sources (0.46 to 9.10 edits per 1{,}000 tokens;
Table~\ref{tab:edit-density}) while mean savings per edit spans $4\times$
(5.9 to 23.3 tokens); both matter (PutnamBench-verified has lower
density than Goedel-Workbook but larger edits, and a larger reduction).
For example, Mathlib's 6{,}695 edits touch 2{,}233 of 5{,}789 files at 5.9
tokens each, whereas miniF2F's 1{,}184 edits touch 308 of 351 files at
23.3 tokens each. This is a descriptive decomposition, not a causal
account, and is consistent with the observed edit mix: tactic replacement and
removal of unused material account for most measured savings.

\paragraph{Which phases matter.}

On a random 500-file Goedel-Workbook slice, disabling tactic
replacement lowers token savings from 6.58\% to 3.86\%; disabling
unused-fact removal or cleanup costs about 1.2 points; disabling
generalization or its dependency filter leaves token savings
essentially unchanged (6.61\%, 6.58\%). On a slice selected for many
\texttt{have} blocks, unused-fact removal dominates (9.08\% $\to$
$-0.23\%$), and disabling generalization slightly \emph{increases}
token savings (9.53\%). Generalization is thus a rare edit family (0.8\%
of accepted edits) whose value we do not establish by compression;
disabling the specificity filter shortens more files (180 vs.\ 167)
without increasing token savings (full table in App.~\ref{app:symbolic}).

\section{Method Details}
\label{app:method}

This section specifies the policy and generalization checks used in
\S\ref{sec:method}. They govern which verified edits the symbolic
teacher prefers; they are distinct from the final correctness check.
Algorithm~\ref{alg:leanpolish} summarizes one pass of \LP{}.

\begin{algorithm}[t]
\caption{One \LP{} pass, with first-success or complete-menu search (\S\ref{sec:method})}
\label{alg:leanpolish}
\small
\textbf{Input:} compiling file $F$; ordered tactic menu $\mathcal{M}$; specificity filter $Q$; mode $\in\{\textsc{first},\textsc{complete}\}$.
\begin{enumerate}[leftmargin=*]
\item Elaborate $F$ once; collect goal-closing tactic sites $s$ (non-structural nodes longer than 3 bytes) with their goals $g_s$.
\item For each site $s$: \emph{first}: try $\mathcal{M}$ in order and stop at the first $c$ that closes $g_s$ and is shorter than $s$;
  \emph{complete}: try all of $\mathcal{M}$, log every outcome, and take the shortest valid, shorter $c$ (ties: menu order).
  Keep $c$ only if $Q(s,c)$ holds (else no edit); if $\mathcal{M}$ yields nothing, try a bounded \texttt{exact?}.
  Record $(s, g_s, c)$ and the failed siblings.
\item Add local-fact generalization, unused-\texttt{have} and unreachable-code candidates.
\item Select non-overlapping candidates greedily by byte savings; re-elaborate the edited file, blame errors on candidates, and re-select (up to 4 rounds).
\item Clean \texttt{<;>} chains with Lean's linters (up to 3 rounds), then re-check the output in a fresh \texttt{lake env lean} process.
\item Return the verified file and its accepted and failed records. Iterating passes until no file changes gives the fixed point.
\end{enumerate}
\end{algorithm}

\paragraph{Specificity filter.}
The filter groups tactics into tiers: \texttt{rfl}; \texttt{ring} and
\texttt{abel} (replaceable only by \texttt{rfl});
\texttt{norm\_num}, \texttt{positivity}, \texttt{norm\_cast},
\texttt{decide} (interchangeable, never replaced by lower tiers);
\texttt{linarith}, \texttt{nlinarith}, \texttt{omega} (peers);
\texttt{field\_simp}, \texttt{gcongr}, \texttt{tauto} (never replaced
by \texttt{simp}). \texttt{trivial}, \texttt{assumption}, and
\texttt{contradiction} are never replaced, and no structural tactic is
replaced by a search-based one (\texttt{simp}, \texttt{tauto},
\texttt{aesop}). Witness tactics (\texttt{exact}, \texttt{rw},
\texttt{apply}, \texttt{refine}, \texttt{use}, \texttt{convert},
\texttt{cases}, \texttt{rcases}, \texttt{obtain}, \texttt{calc},
\texttt{induction}, \texttt{match}, \texttt{constructor},
\texttt{left}, \texttt{right}, \texttt{conv}, \ldots) are never replaced
unless the original is a multi-step block (\texttt{;} or \texttt{<;>}).
This tier structure is distinct from the \emph{search} order of
\S\ref{sec:method}: the search proposes the first successful candidate,
and the filter then accepts or rejects that single proposal. Additional
guards reject rewrites of non-\texttt{Prop} goals, goals closed through
\texttt{sorryAx}, and \texttt{exact?} suggestions whose head constant
is not from the imported library (e.g.\ the theorem being proved).

\paragraph{Dependency filter for generalization.}
For a group of $N$ near-duplicate \texttt{have} blocks with proof terms
$p_1,\dots,p_N$, let $U$ be the set of local free variables occurring in
any $p_i$, and let $q$ be the anti-unified proof after abstracting the
differing subterms into parameters (via
\texttt{mkForallFVars}/\texttt{mkLambdaFVars}, checked with
\texttt{Meta.check} and \texttt{isDefEq}). The merge is rejected if
$|\mathrm{FV}(q)| \ge |U|$ (logged as
\texttt{g3\_no\_generalization\_$a$ge$b$}). This filter is a heuristic;
we make no information-theoretic claim for it, and a count of
parameters or free variables alone does not determine whether a merge
shortens the proof. Table~\ref{tab:g3} gives
the rule's behavior on a stratified sample.

\begin{table}[h]
\caption{Local fact generalization on a stratified random sample
(seed 42).}
\label{tab:g3}
\centering
\small
\begin{tabular}{lrr}
\toprule
& Goedel-Workbook & Mathlib v4.21.0 \\
\midrule
Files sampled & 1{,}500 & 250 \\
Candidate groups discovered & 86 & 49 \\
Reached the dependency filter & 16 & 1 \\
Rejected by dependency filter & 6 & 0 \\
Applied & 10 & 1 \\
\bottomrule
\end{tabular}
\end{table}

\paragraph{Complexity and runtime.}
Per file, tactic replacement makes $O(T)$ bounded Lean calls for $T$
leaf tactics; unused-fact removal builds a usage graph linear in the
number of local hypotheses and needs at most one check per block;
generalization compares blocks only within signature buckets. Each Lean process loads Mathlib once and processes files sequentially;
a Python orchestrator parallelizes across processes. The full
29{,}750-file Goedel-Workbook run takes about one hour on a 96-core
machine; 309 files (1.0\%) fail (timeout or error) and are counted as
unshortened.

\section{Symbolic Results: Full Tables}
\label{app:symbolic}

These tables give the alternative size metrics and phase ablations
behind \S\ref{sec:ablation-summary} and App.~\ref{app:symbolic-main}.
They separate the source of compression from the mere presence of a
phase in the pipeline.

\begin{table}[h]
\caption{Whole-corpus reduction (\%) in bytes (B), tokens (T), and
non-blank lines (L); files not shortened count as 0\%.}
\centering
\small
\begin{tabular}{lrrrrrr}
\toprule
& \multicolumn{3}{c}{\LP{}} & \multicolumn{3}{c}{\texttt{linter.unusedTactic}} \\
\cmidrule(lr){2-4}\cmidrule(lr){5-7}
Corpus & B & T & L & B & T & L \\
\midrule
miniF2F (Goedel-V2 verified) & 19.83 & 19.72 & 21.59 & --- & 0.002 & --- \\
PutnamBench (Goedel-V2 verified) & 6.22 & 6.14 & 7.29 & --- & 0.000 & --- \\
Mathlib v4.21.0 & 0.24 & 0.27 & 0.18 & 0.000 & 0.000 & 0.000 \\
Goedel-Workbook & 10.66 & 5.48 & 12.05 & 0.018 & 0.010 & 0.000 \\
PutnamBench (Goedel sample) & 8.82 & 8.14 & 11.39 & 0.013 & 0.007 & 0.000 \\
Putnam 2025 / AxiomProver & 1.21 & 1.31 & 1.20 & 0.000 & 0.000 & 0.000 \\
\bottomrule
\end{tabular}
\end{table}

\paragraph{Linter reproduction.}
We enable \texttt{linter.unusedTactic}, disable
\texttt{linter.unreachableTactic} and \texttt{linter.unnecessarySeqFocus},
fix \texttt{maxHeartbeats} at 800{,}000, widen detected ranges to
trailing separators, splice them out, and re-elaborate. ProofOptimizer
reports its linter figures on 194 / 75 Goedel-Prover-V2 proofs under
Mathlib 4.19; we apply the configuration to 351 / 19 Goedel-Prover-V2 proofs
under v4.21; we have not isolated whether the proofs or the Mathlib
version explain why the linter rarely fires here.

\begin{table}[h]
\caption{Phase ablation on Goedel-Workbook. (a) Random 500-file slice;
(b) 500 files selected for many \texttt{have} blocks. Reduction in \%;
Gen = applied generalizations; Filt = generalizations rejected by
the dependency filter. Identical aggregate rows indicate no measured change; phase
interactions can offset individual edits.}
\label{tab:ablation}
\centering
\small
\begin{tabular}{lrrrrrr}
\toprule
Configuration & Shorter & B & T & L & Gen & Filt \\
\midrule
\multicolumn{7}{l}{\emph{(a) random slice}} \\
Full pipeline & 167 & 11.75 & 6.58 & 9.90 & 2 & 3 \\
-- tactic replacement & 108 & 1.94 & 3.86 & 6.31 & 2 & 3 \\
-- generalization & 166 & 11.74 & 6.61 & 9.88 & 0 & 0 \\
-- unused-fact removal & 151 & 11.31 & 5.40 & 8.66 & 2 & 3 \\
-- cleanup & 131 & 11.06 & 5.34 & 8.09 & 2 & 3 \\
-- specificity filter & 180 & 11.75 & 6.58 & 9.81 & 2 & 3 \\
-- dependency filter & 167 & 11.75 & 6.58 & 9.90 & 2 & 0 \\
\midrule
\multicolumn{7}{l}{\emph{(b) \texttt{have}-rich slice}} \\
Full pipeline & 245 & 5.92 & 9.08 & 10.64 & 57 & 15 \\
-- tactic replacement & 240 & 5.75 & 9.07 & 10.64 & 60 & 14 \\
-- generalization & 240 & 5.70 & 9.53 & 10.31 & 0 & 0 \\
-- unused-fact removal & 44 & 0.51 & $-$0.23 & 0.60 & 57 & 2 \\
-- cleanup & 245 & 5.92 & 9.08 & 10.64 & 57 & 15 \\
-- specificity filter & 254 & 5.44 & 8.67 & 9.83 & 56 & 16 \\
-- dependency filter & 245 & 5.92 & 9.08 & 10.64 & 57 & 0 \\
\bottomrule
\end{tabular}
\end{table}

\section{Frontier-Prover Details}
\label{app:frontier}

This section supports the scope of the frontier-source comparison in
\S\ref{sec:symbolic}. It records the shared-toolchain subset and
failures so that source differences are not mistaken for a controlled
comparison of prover quality.

\begin{table}[h]
\caption{Seed-Prover~1.5 vs.\ AxiomProver Putnam~2025 solutions under a
shared Lean/Mathlib 4.22 toolchain, on the eight problems where both
releases compile. Tokens removed / original tokens.}
\centering
\small
\begin{tabular}{lrr}
\toprule
Problem & Seed-Prover 1.5 & AxiomProver \\
\midrule
A1 & 371 / 9{,}105 (4.1\%) & 118 / 7{,}067 (1.7\%) \\
A2 & 0 / 6{,}115 (0.0\%) & 72 / 5{,}385 (1.3\%) \\
B1 & 0 / 12{,}039 (0.0\%) & 115 / 15{,}590 (0.7\%) \\
B2 & 1{,}702 / 27{,}478 (6.2\%) & 64 / 5{,}502 (1.2\%) \\
B3 & 0 / 6{,}063 (0.0\%) & 76 / 3{,}552 (2.1\%) \\
B4 & 230 / 8{,}503 (2.7\%) & 393 / 13{,}360 (2.9\%) \\
B5 & 0 / 34{,}371 (0.0\%) & 128 / 16{,}575 (0.8\%) \\
B6 & 1{,}500 / 35{,}816 (4.2\%) & 0 / 13{,}076 (0.0\%) \\
\midrule
Total & 3{,}803 / 139{,}490 (2.7\%) & 966 / 80{,}107 (1.2\%) \\
\bottomrule
\end{tabular}
\end{table}

Inputs are the unmodified official releases (file hashes recorded in
the code repository). The 4.22 port adds a single
\texttt{maxHeartbeats} option to the optimizer (not to the inputs) and
raises the per-file budget to 7{,}200\,s; the algorithm, phases, filters, and
tokenizer are unchanged. Under 4.22, 11/11 Seed-Prover files and 8/12
AxiomProver files compile; where a problem was run under both
toolchains, savings agree exactly on three of five problems and closely
on the other two. Seed-Prover B2 completed only under 4.22 (it exceeded a
14{,}400\,s budget under 4.21). Aristotle's IMO~2025 release pins Lean
4.20-rc5; P3 and P5 are processed end to end (0.3\%, 0.5\%), one file does
not compile under our toolchain, and P4 (103\,KB) triggers a
deterministic out-of-range indexing error in the optimizer.

\section{Learning Experiment Details}
\label{app:learning}

This section separates the learning tasks in \S\ref{sec:selection}
from the whole-file tests in \S\ref{sec:use}. Conditional editing
predicts a span at a teacher-selected site; complete-pool ranking
selects among logged alternatives; whole-proof rewriting generates an
entire proof. Their success rates have different denominators.

\begin{algorithm}[t]
\caption{Verified hybrid compression (\S\ref{sec:hybrid})}
\label{alg:hybrid}
\small
\textbf{Input:} compiling file $F$, editor $M$, number of proposals $k$.
\begin{enumerate}
\item Run the verified symbolic pass: $F_1\gets\textsc{LeanPolish}(F)$.
\item Extract tactic sites $S$ and their goal states from $F_1$; set $E=\emptyset$.
\item For each $s\in S$, generate $k$ proposals and retain only shorter spans, forming $C_s$.
\item Verify each splice independently:
  $V_s=\{c\in C_s:\textsc{Compiles}(F_1[s\mapsto c])\}$.
  If $V_s\ne\emptyset$, add a candidate with maximum savings and its site to $E$.
\item Apply a compatible, non-overlapping subset of $E$ and verify the whole file.
  If joint verification fails, reduce the subset; use $F_1$ if no subset succeeds.
\item Return the jointly verified file and measure its savings relative to $F$.
\end{enumerate}
\end{algorithm}

\paragraph{Prompt.} \texttt{You are editing a Lean 4 proof. Return only
a shorter replacement that preserves the goal. Return <DELETE> if the
fragment should be removed.} followed by \texttt{[GOAL]} (the goal
state, or ``(no local goal)''), \texttt{[LOCAL CONTEXT]} (surrounding
source), \texttt{[ORIGINAL FRAGMENT]}, and \texttt{[REPLACEMENT]}.

\paragraph{Local-editor training and decoding.} LoRA rank 32, $\alpha=64$, dropout 0.05 on all
linear layers; 2 epochs; learning rate $10^{-4}$ with cosine schedule
and 3\% warmup; weight decay 0.01; effective batch 64; seed 42; BF16 on
one H100. Decoding: greedy, plus 4 samples at temperature 0.6, top-$p$
0.95 for best-of-4 analyses.

\paragraph{Verification.} Candidates are first screened in a persistent
Lean REPL; every candidate counted as a success then receives an
individual verdict from a fresh \texttt{lake env lean} compilation of the
spliced file, including reference-identical outputs. The original local-editor verification batch contains
2{,}086 distinct candidate splices with an exact verdict,
with no conflicting verdicts and full agreement with the screen.

\paragraph{Selection among sampled outputs.} Pooled verified savings
over the three held-out sources when choosing among four SFT samples:
greedy 29{,}166 tokens; random choice 28{,}832 (expectation); learned
ranker 28{,}720; shortest string 28{,}354; verifier oracle 29{,}250.
The ranker selects fewer savings than random choice.

\paragraph{Complete-pool ranker, further splits.} On the in-distribution
Goedel-Workbook validation pools the ranker reaches 80.7\% top-1 best
(no goal: 68.8\%; validity AUROC 0.994); on a 3{,}000-state sample of
human-written Mathlib proofs it reaches 53.8\% (no goal: 41.3\%).
On the Mathlib sample, first-success verified search (menu order, stop at
the first acceptable candidate) reaches 91.4\%. Ranker: DeepSeek-Prover-V2-7B
with LoRA ($r{=}16$), two pointwise heads (valid, best), trained for one
epoch on 80{,}343 candidates, whole sites subsampled (seed 42) from
191{,}321 candidate rows of 13{,}290 Goedel training states.

\paragraph{DPO.} One epoch on 16{,}839 same-attempt pairs (chosen =
accepted edit, rejected = failed sibling), initialized from the DeepSeek
SFT model (LoRA $r=16$, $\alpha=32$, learning rate $10^{-5}$,
$\beta=0.1$), same verification; results are in Table~\ref{tab:sft}.
Because DPO pairs inherit the search-order structure of
\S\ref{sec:ranking}, we treat DPO as a check of the preference
interface, not as an improvement method.

\paragraph{Whole-proof training and self-training.}
The whole-proof rewriter uses 7{,}708 original/output Goedel-Workbook
pairs, including 11\% unchanged pairs, and 67 GPU-minutes of training;
evaluation uses $k=16$ except where Table~\ref{tab:hybrid} states
otherwise. These experiments use one checkpoint per configuration.
The prompt is ``Rewrite this Lean 4 proof to be shorter while remaining
correct. Output the full file.'' followed by the file. Pairs come only
from files in the seed-42 file-level SFT training split (validation files
and goal-overlap drops excluded), are rebuilt byte-exactly from the edit
shards (360 failing reconstruction checks dropped), and fit an
8{,}192-token budget (max 3{,}868). Training uses LoRA on
DeepSeek-Prover-V2-7B ($r{=}32$, $\alpha{=}64$, dropout 0.05, attention
and MLP projections), learning rate $10^{-4}$ (cosine, 3\% warm-up), two
epochs, loss on the output only, 467 steps. Sampling uses $T{=}0.7$,
top-$p$ 0.95, seed 42; the sample rate is the fraction of all samples
that yield a verified, strictly shorter file.
For local-editor self-training, 410 shorter proposals from 5{,}472
sites compile, leaving 372 after deduplication. Only two are deletions;
147 (40\%) violate the specificity filter. The new edits save 0.55\%
of input tokens under the comment-counting counter. We compare continued
training on all new edits with training on filter-compliant non-deletions,
each mixed with a fixed random replay sample (seed 42) of SFT training
rows equal in size to the naive new-edit set (744 and 595 rows in total).
Both continue from the round-1 adapter for one epoch with the round-1
recipe (learning rate $10^{-4}$ cosine, 3\% warm-up, weight decay 0.01,
effective batch 64, maximum length 2{,}048). The update comprises only 10--12
optimizer steps; a from-base retrain also finds no gain. Filter compliance
is 96\% at teacher-selected sites, 32\% on the tested non-teacher sites,
and 34\% after naive self-training. These observations motivate the
limited interpretation in \S\ref{sec:selfimprove}.

\section{Leakage Audit}
\label{app:leakage}

This audit supports the split in \S\ref{sec:selection}. Hashes compare
whitespace-normalized goal strings, not semantic equivalence; low exact
overlap does not rule out related statements or overlap in model
pretraining. The PutnamBench-sample rows below describe the release but
are not a training source for the local editors. Non-goal deletion rows
are separated by source rather than by placeholder hashes.

\begin{table}[h]
\caption{Goal-hash overlap between training and held-out sources
(unique whitespace-normalized goal hashes; Jaccard is a fraction).}
\centering
\small
\begin{tabular}{lrrrr}
\toprule
Train $\cap$ held-out & $|$Train$|$ & $|$Held-out$|$ & $|\cap|$ & Jaccard \\
\midrule
Goedel $\cap$ miniF2F & 5{,}193 & 696 & 9 & 0.0015 \\
Goedel $\cap$ PutnamBench-verified & 5{,}193 & 38 & 1 & 0.0002 \\
Goedel $\cap$ Putnam 2025 & 5{,}193 & 55 & 1 & 0.0002 \\
Mathlib $\cap$ miniF2F & 6{,}490 & 696 & 2 & 0.0003 \\
Mathlib $\cap$ PutnamBench-verified & 6{,}490 & 38 & 1 & 0.0002 \\
Mathlib $\cap$ Putnam 2025 & 6{,}490 & 55 & 2 & 0.0003 \\
PutnamBench sample $\cap$ miniF2F & 561 & 696 & 4 & 0.0032 \\
PutnamBench sample $\cap$ PutnamBench-verified & 561 & 38 & 8 & 0.0135 \\
PutnamBench sample $\cap$ Putnam 2025 & 561 & 55 & 1 & 0.0016 \\
\bottomrule
\end{tabular}
\end{table}

\section{Release Schema}
\label{app:schema}

The schema implements the local supervision record in \S\ref{sec:method}.
Consumers should distinguish original first-success siblings from the
complete-menu outcomes illustrated in App.~\ref{app:usage}.

Accepted rows (\texttt{schema\_version} 2) expose the following fields
where applicable: \texttt{original},
\texttt{replacement}, \texttt{goal\_state}, \texttt{goal\_pretty},
\texttt{goal\_type}, \texttt{type} (edit family), \texttt{kind} (Lean
syntax kind), \texttt{start\_byte}, \texttt{end\_byte}, \texttt{line},
byte/token/line counts of both spans, \texttt{edit\_width} (UTF-8 byte
difference), \texttt{savings}, \texttt{term\_size}, \texttt{context},
\texttt{file}, \texttt{corpus}, \texttt{attempt\_id},
\texttt{rank\_in\_attempt}, \texttt{outcome}, \texttt{failed\_tactics},
\texttt{failed\_attempts} (tactic, error, wall time), and provenance
(optimizer revision as \texttt{git\_sha} or \texttt{commit\_sha}, \texttt{mathlib\_rev},
\texttt{content\_sha256}). Rejected rows share the same fields plus
\texttt{err\_msg} and \texttt{wall\_ms}, with outcome
\texttt{rejected\_attempt}; finer failure reasons (kernel error,
unsolved goals, timeout, parse error) are derived from \texttt{err\_msg}
and are not a separate label. For non-tactic rows \texttt{goal\_type}
holds a family placeholder and \texttt{goal\_state} is empty; these
fields must not be used for goal-hash deduplication of such rows. For
strict compression training we recommend keeping rows with positive
\texttt{edit\_width} (33{,}360 of 33{,}402).

\section{Data Notes}
\label{app:errata}

These notes qualify the release counts in Table~\ref{tab:corpora},
byte-exact replay, and the comparison denominators in
Table~\ref{tab:hybrid}.

(i)~\emph{Hashes.} In the miniF2F and PutnamBench-verified shards,
\texttt{content\_sha256} is the hash of the working file when the row
was emitted, which differs from the input file once earlier edits in the
same file have been applied; edit spans are byte-exact against the input
files, and the replay verifier ignores this field for these shards.
(ii)~\emph{License.} The release is distributed under Apache-2.0 (dataset
card and Croissant manifest); upstream licenses are listed per source.
(iii)~\emph{Units.} \texttt{edit\_width} is in UTF-8 bytes, not
characters. (iv)~\emph{Linter-baseline files.} The Goedel-Workbook and
PutnamBench-sample shards contain 92 files named \texttt{*\_linter.lean}
(133 edits), which are outputs of the linter baseline rather than
optimizer inputs; they are listed in the release. They are not used as optimizer inputs
in any evaluation; the one such file in the random miniF2F-100 subset is
removed, so every miniF2F row of Table~\ref{tab:hybrid} uses the same 99
files. (v)~\emph{Denominators.} All reductions are computed over
exact input-file lists, released in the code repository. (vi)~\emph{Machine
dependence.} \LP{} outcomes depend on machine speed through
per-candidate and per-file time budgets: a clean rerun of the release
binary shortens PutnamBench-verified by 11.9\% instead of 6.1\% (the
release run hit budgets on several files), and a single pass is not a
fixed point. We report both runs. The local-editor hybrid block starts from the
release run; composed-pipeline rows explicitly identify clean reruns.

\clearpage
\section{Using LeanPolish and the Release}
\label{app:usage}

These recipes connect the resource in \S\ref{sec:method} to the local
learning and whole-file experiments in \S\ref{sec:selection}--\ref{sec:use}.
The pool example makes the first-success shortcut concrete.

\paragraph{Recipes.} All commands are in the code repository (its README
lists every script and the table it reproduces).
\begin{enumerate}\raggedright\itemsep0pt
\item \emph{Build} (Lean~4.21.0, Mathlib v4.21.0):
  \texttt{lake exe cache get \&\& lake build LeanPolish}.
\item \emph{Compress a folder of proofs}:
  \texttt{python3 leanpolish.py -{}-batch DIR -{}-workers 6 -{}-threads 4 -{}-timeout 1800} writes \texttt{*\_shortened.lean}, a per-file report,
  and the accepted and failed edits; use the independent verifier below
for byte-exact replay.
\item \emph{Fixed point}: re-run step 2 on the shortened outputs until no
  file changes (\texttt{complete\_menu/fixpoint.py} automates it; 4--8
  rounds in our corpora).
\item \emph{Complete candidate pools}: add \texttt{-{}-complete-menu}; each
  site then lists all 15 menu entries, including explicit length skips,
  (format on the dataset card; example in Table~\ref{tab:poolex}).
\item \emph{Verify any edit independently}: \texttt{verify\_pair.py ROWS.jsonl -{}-project-root DIR} splices each edit into its source
  file and compiles it in a fresh process.
\item \emph{Train and run a local editor}: \texttt{build\_sft\_data.py}
  $\to$ \texttt{train\_sft\_lora.py}; the verified hybrid of
  Algorithm~\ref{alg:hybrid} is \texttt{beyond\_teacher/scripts/pipeline.sh};
  both LoRA adapters are released.
\end{enumerate}

\paragraph{What a complete pool records.} Table~\ref{tab:poolex} shows one
held-out state. First-success search stops at \texttt{linarith}, so the
first-success release would store the seven earlier failures as its only
negatives and never test \texttt{omega}; the complete pool shows four valid
candidates, a character-shorter best one, and a label for every entry.
The valid one-word tactics all have the same token count: this example
illustrates a ranking distinction, not an additional token saving.

\begin{table}[H]
\caption{One complete pool (PutnamBench \texttt{putnam\_1975\_a1}). Goal
$\uparrow n = (\uparrow a^2 + \uparrow a)/2 + (\uparrow b^2 + \uparrow b)/2$;
original span \texttt{simpa [add\_assoc] using h$_1$}. Entries in menu
order; outcomes as released.}
\label{tab:poolex}
\centering
\small
\begin{tabular}{p{0.36\linewidth}lp{0.36\linewidth}}
\toprule
Candidate & Outcome & Note \\
\midrule
\raggedright\texttt{rfl}, \texttt{ring}, \texttt{abel}, \texttt{norm\_num}, \texttt{norm\_cast}, \texttt{positivity}, \texttt{decide} & kernel failure & \raggedright tried and failed: the only negatives a first-success release stores \tabularnewline
\texttt{linarith} & valid, not chosen & first-success winner (8 characters) \\
\texttt{omega} & \textbf{chosen} & shortest valid candidate (5 characters) \\
\raggedright\texttt{field\_simp}, \texttt{contradiction}, \texttt{ext}, \texttt{simp?} & kernel failure & \raggedright never tried by first-success search \tabularnewline
\texttt{gcongr}, \texttt{tauto} & valid, not chosen & never tried by first-success search \\
\bottomrule
\end{tabular}
\end{table}

\clearpage
\section{Worked Examples}
\label{app:examples}

The examples illustrate the proposal families of \S\ref{sec:method}
and the policy distinction tested in \S\ref{sec:training-utility}.
Ellipses mark abridged excerpts; the displays are not standalone Lean
files, and numerical savings refer to the complete recorded edits.

\paragraph{Symbolic edits.} The excerpts illustrate, in order: a goal closed by \texttt{simp} that is definitional, a
speculative tactic cascade that \texttt{ring} alone closes, and
combinator arms that run after the goal is already closed.

\diffheader{Tactic replacement (\texttt{simp} $\to$ \texttt{rfl})}{lean\_workbook\_10208.lean}
\begin{lstlisting}[style=leandiff]
  | zero =>
-   simp
+   rfl
\end{lstlisting}

\diffheader{Cascade collapsed to \texttt{ring}}{lean\_workbook\_1001.lean}
\begin{lstlisting}[style=leandiff]
- := by
-   simp only [sq, mul_add, mul_comm, mul_left_comm, mul_assoc,
-              add_assoc, add_left_comm]
-   ring
-   <;> simp [Complex.I_sq]
-   <;> ring
+ := by ring
\end{lstlisting}

\diffheader{Unreachable-code cleanup}{lean\_workbook\_plus\_61403.lean}
\begin{lstlisting}[style=leandiff]
  ring_nf
- <;> simp_all only [sub_eq_add_neg]
- <;> ring_nf
- <;> simp_all only [sub_eq_add_neg]
- ... (22 more unreachable arms)
\end{lstlisting}

\paragraph{Verified neural edits (hybrid, \S\ref{sec:use}).} The full versions of all three edits are
accepted by Lean. The first two also pass \LP{}'s specificity filter
(pruning unneeded hints; dropping a type the elaborator infers, an edit
outside the symbolic menu). The last does not: it replaces an explicit
proof term by automation, illustrating the policy relaxation in the trained hybrid.

\diffheader{Hint pruning (compliant; $-37$ tokens)}{PutnamBench \texttt{putnam\_1993\_a2}}
\begin{lstlisting}[style=leandiff]
- nlinarith [sq_pos_of_ne_zero (xnonzero k), ... (4 hints)]
+ linarith
\end{lstlisting}

\diffheader{Redundant type dropped (compliant; $-30$ tokens)}{miniF2F}
\begin{lstlisting}[style=leandiff]
- have h(*@$_4$@*) : f (-2 + 3) = 3 * (-2 : (*@$\mathbb{R}$@*)) ^ 2 + 7 * (-2 : (*@$\mathbb{R}$@*)) + 4 := h(*@$_0$@*) (-2)
+ have h(*@$_4$@*) := h(*@$_0$@*) (-2)
\end{lstlisting}

\diffheader{Term $\to$ \texttt{tauto} (filter-rejected; $-27$ tokens)}{AxiomProver}
\begin{lstlisting}[style=leandiff]
- exact (*@$\langle$@*)fun (*@$\langle$@*)h1, h2(*@$\rangle$@*) => (*@$\langle$@*)h2, h1(*@$\rangle$@*), fun (*@$\langle$@*)h1, h2(*@$\rangle$@*) => (*@$\langle$@*)h2, h1(*@$\rangle\rangle$@*)
+ tauto
\end{lstlisting}

\end{document}